\pdfoutput=1

\documentclass[11pt]{article}

\usepackage[final]{acl}

\usepackage{times}
\usepackage{latexsym}

\usepackage[T1]{fontenc}
\usepackage[utf8]{inputenc}
\usepackage{CJKutf8}

\usepackage{microtype}

\usepackage{inconsolata}

\usepackage{graphicx}

\usepackage{array}
\usepackage{algorithm}
\usepackage{algpseudocode}

\usepackage{pgfplots}
\usepackage{pgfplotstable} 
\pgfplotsset{compat=1.16} 
\usepgfplotslibrary{statistics}

\title{Data Quality Issues in Multilingual Speech Datasets: The Need for Sociolinguistic Awareness and Proactive Language Planning}

\author{Mingfei Lau \quad Qian Chen \quad Yeming Fang \quad Tingting Xu \quad Tongzhou Chen \quad Pavel Golik \\
  Google \\
  \texttt{\{laufei, allenqc, yeming, xutt, tongzhou, golik\}@google.com}}

\DeclareUnicodeCharacter{030D}{\mbox{\normalfont\textsuperscript{|}}}
\begin{document}
\maketitle
\begin{abstract}
Our quality audit for three widely used public multilingual speech datasets\textemdash Mozilla Common Voice 17.0, FLEURS, and VoxPopuli\textemdash shows that in some languages, these datasets suffer from significant quality issues, which may obfuscate downstream evaluation results while creating an illusion of success. We divide these quality issues into two categories: micro-level and macro-level. We find that macro-level issues are more prevalent in less institutionalized, often under-resourced languages. We provide a case analysis of Taiwanese Southern Min (\texttt{nan\_tw}) that highlights the need for proactive language planning (e.g. orthography prescriptions, dialect boundary definition) and enhanced data quality control in the dataset creation process. We conclude by proposing guidelines and recommendations to mitigate these issues in future dataset development, emphasizing the importance of sociolinguistic awareness and language planning principles. Furthermore, we encourage research into how this creation process itself can be leveraged as a tool for community-led language planning and revitalization.

\end{abstract}

\section{Introduction}

The emergence of massively multilingual speech datasets has significantly advanced the performance of various speech technologies in recent years, particularly for low-resource languages. These datasets are crucial for training and evaluating state-of-the-art ASR models like Whisper \cite{radford2023robust}, Google USM \cite{zhang2023google}, SeamlessM4T \cite{barrault2023seamlessm4t}, MMS \cite{pratap2024scaling}, and Gemini \cite{team2024gemini}, and also enable advances in cross-lingual speech representation learning \citep{babu2021xls, conneau2020unsupervised} and downstream applications like multilingual speech generation and understanding \citep{NEURIPS2023_2d8911db, rubenstein2023audiopalm}. However, despite their growing importance, the quality of these datasets remains surprisingly under-researched.

Prior work on data collection and curation \citep{penedo2023refinedweb, goyal2022flores, kreutzer2022quality} has acknowledged the generally lower quality of web-scraped data, but these efforts primarily focused on text. Similarly, research on ASR data augmentation for low-resource languages \citep{casanova2022asr, bartelds2023making, tsoukala2023asr} has not addressed data quality issues adequately. The community-driven Mozilla Common Voice project \cite{ardila2019common}, for example, lacks well-documented quality control processes for its diverse text sources (Wikipedia and volunteer contributions) and subsequent audio recordings, which makes the quality and reliability largely unknown.

Inspired by \citet{kreutzer2022quality}'s audit methodology for text datasets, we conduct a thorough quality assessment of three widely used multilingual speech datasets: Mozilla Common Voice 17.0 (MCV17, \citealt{ardila2019common}), FLEURS~\cite{conneau2023fleurs}, and VoxPopuli~\cite{wang2021voxpopuli}. Our investigation employs both quantitative and qualitative methods. We calculate metrics including Signal-to-Noise Ratio (SNR), Voice Activity Detection (VAD), median utterance duration, and median word count for each language subset. Qualitatively, we asked native speaker volunteers, covering around 40 languages detailed in Table~\ref{tab:qual} in Appendix, to review 100 randomly sampled sentences (text and audio) for coherence, audio-text alignment, dialect, topic domain, and language classification from each language subset of the datasets.

Our analysis reveals serious data quality issues, particularly in low-resource, less-institutionalized languages (Sections \ref{sec:micro} and \ref{sec:macro}). As one example, the \texttt{nan\_tw} (Taiwanese Southern Min) subset in MCV17 showcases a multitude of these issues, rendering it nearly unusable without significant cleaning or restructuring (Section \ref{sec:extreme}). We find a strong positive correlation between a language's institutionalization status and its dataset quality, a crucial factor often overlooked in ASR research. We discuss the impact of these issues on downstream research and applications, and propose mitigation guidelines (Section \ref{sec:impact}). We draw conclusions and suggest future work in Section~\ref{sec:conclusions}.

\section{Datasets}

In this work, we study three multilingual datasets: Mozilla Common Voice 17.0 (MCV17), FLEURS, and VoxPopuli. 

\begin{itemize}

\item Mozilla Common Voice~\cite{ardila2019common} is a community-driven project proposed by Mozilla. Sentence sourcing, recording and reviewing are all contributed by volunteers. Version 17 supports 124 locales.

\item FLEURS~\cite{conneau2023fleurs} uses the text from the machine translation corpus FLoRes-101~\cite{goyal2022flores}. Sentences are extracted from the English Wikipedia and translated to 101 languages by professional translators. Each sentence is recorded by 3 native speakers and  invalid recordings are discarded.

\item VoxPopuli~\cite{wang2021voxpopuli} uses speech and transcriptions from European Parliament event recordings. The paired data contains 16 European languages.

\end{itemize}


\section{Micro-Level Issues} \label{sec:micro}

Micro-level issues in ASR datasets typically stem from inadequate quality control, supervision, or management during data collection and curation. They are often detectable via automatic metrics and can be mitigated programmatically. Micro-level issues are language-agnostic and may exist in any language subsets. 

\subsection{Extremely Short Duration}
\label{sec:shortduration}
In VoxPopuli the median utterance duration ranges from around 6 to 13 seconds and FLEURS ranges from 9 to 24 seconds. However, MCV17 displays a concerning trend of extremely short utterances. Detailed plots are shown in Appendix~\ref{sec:medianduration}. In MCV17, 35 languages have median utterance duration under 4 seconds. And for most languages, over 99\% of the utterances are under 10 seconds. We also find some extreme cases such as \texttt{nan\_tw} (Taiwanese Southern Min), \texttt{sr} (Serbian), and \texttt{br} (Breton), whose median utterance durations are below 3 seconds, and 99\% of the utterances are below 7 seconds. We manually inspect those language subsets and discover that the text prompts of these languages are mostly short phrases or isolated words, as shown in Figure~\ref{fig:mcv17len}. Previous research has discovered that ASR and TTS models may fail to generalize to inputs of lengths not seen in training~\citep{narayanan19,varis21,chiu2021rnn,mengke24}. Without this insight, a model trained on MCV17 \texttt{nan\_tw} might fail on long-form tasks like video subtitling. While mitigation is possible at the level of model architecture (e.g. more flexible attention variants), we believe this needs to be fixed at the dataset design level, as Common Voice does not enforce a strict length requirement for contributed text prompts, leading to the ingestion of low quality inputs.

\subsection{Low Proportion of Speech}
To assess the proportion of actual speech content within the audio data, we employed a neural model to classify speech and non-speech segments in all three datasets. While all 14 languages in VoxPopuli have at least 89\% speech, we identified several languages in other datasets with significantly lower proportions, detailed in Appendix~\ref{sec:speechsilence}. Specifically, Basaa (\texttt{bas}), Zaza (\texttt{zza}), Serbian (\texttt{sr}) in MCV17, and Danish (\texttt{da\_dk}) in FLEURS exhibiting less than 50\% speech content,. We manually inspected these languages and found two primary causes: 
\begin{itemize}
    \item The speaker's voice is too distant from the microphone, resulting in poor audio capture (e.g. the \texttt{da\_dk} training set in FLEURS).
    \item The short text prompt issue described in Section~\ref{sec:shortduration} reduces the proportion of actual speech in the utterance, resulting in an increased proportion of silence between the start and stop of the recording process, as contributors interacted with their recording devices.
\end{itemize}
Similar issues have also been reported by the public\footnote{A large number of incorrect audio samples on FLEURS reported in April 2023
\newline
\href{https://huggingface.co/datasets/google/fleurs/discussions/16}{https://huggingface.co/datasets/google/fleurs/discussions/16}}
which further validates our findings.
This issue combined with extremely short utterance durations severely limits the amount of usable speech data. For instance, while \texttt{nan\_tw} in MCV17 contains 21 hours of audio, only 48.3\% constitutes actual speech, resulting in merely 10 hours of usable data.
While in general this micro-level issue affects TTS more than ASR, using the raw audio duration e.g. for re-balancing the amounts of training data per language can lead to unexpected results.  This can be mitigated by using the number of transcribed words (or tokens) as the most fundamental unit for quantifying speech data.

\subsection{Imbalanced Topic Domains}

Assessing topic domain balance is challenging to quantify programmatically. Therefore, we adopted a qualitative approach, sampling text prompts and corresponding audio data from each dataset. We asked native speakers to evaluate whether these sentences were representative of typical, everyday conversations—a common target domain for general-purpose ASR systems. However, we acknowledge that other domains (e.g., news broadcasts, academic lectures) might be relevant for specific applications. This investigation revealed significant topic domain imbalances in several languages within both MCV17 and FLEURS, particularly concerning the assumption of everyday conversation as the target. In FLEURS, the reliance on text prompts from the FLoRes-101 dataset \cite{goyal2022flores}, ultimately sourced from Wikipedia, leads to a dominance of formal, literary, and encyclopedic sentences.

We also identified an issue in MCV17, where a number of sentences across multiple languages exhibit high repetitiveness, often lacking meaningful content and suggesting machine generation using fixed templates (see Table~\ref{tab:repeat} in Appendix). It's important to acknowledge that these datasets are built on the valuable contributions of volunteers. Given the community-driven nature of text prompt creation in Common Voice, it is possible that these repetitive sentences were introduced due to varying interpretations of the guidelines or a lack of awareness about their potential impact on model training. This highlights the need for clearer guidelines and more robust review processes within community-driven projects.

The problem of domain mismatch has been discussed in the context of monolingual datasets in \cite{likhomanenko21}. In multilingual datasets, a poor coordination in topic selection across languages might lead to incorrect interpretation of ASR and TTS model quality. While there is no easy mitigation, at the very least this information should be exposed in machine-readable metadata.

\subsection{Lack of Speaker Diversity}
\label{speakerdiversity}
The final micro-level issue we identified is a lack of speaker diversity. We calculated the average recording time per speaker for each language in MCV17, as detailed in Tables~\ref{tab:hoursperspeaker} and~\ref{tab:speakers} in the Appendix. Notably, Macedonian (\texttt{mk}) exhibits an average of 1.20 hours of audio per speaker, with only 19 unique speakers in total, indicating a high concentration of data from a limited number of contributors. Data of some languages such as Zulu (\texttt{zu}), Northern Sotho (\texttt{nso}), Haitian Creole (\texttt{ht}) consist of recordings from only a single speaker.

This introduces a high risk of overfitting to the speakers' age, gender and dialects, as well as a risk of increasing bias in downstream models \cite{koenecke20}. Whenever the speaker statistics are available as metadata, dataset users can focus on the subset of languages with diverse data, which produce reliable results.


\section{Macro-Level Issues} \label{sec:macro}

Macro-level issues often arise from overlooking a language's sociolinguistic context during dataset design. These issues are typically not detectable through automatic metrics, requiring manual inspection and linguistic expertise for diagnosis. They are particularly prevalent in less-institutionalized languages, often characterized by complex phenomena like digraphia or diglossia. Our analysis of VoxPopuli, which primarily includes more-institutionalized languages, further supports this claim, as it revealed no such macro-level issues.

\subsection{Unspecified Writing System in Digraphic Languages}

Digraphia is a sociolinguistic phenomenon where a single language is written using multiple writing systems \cite{dale1980digraphia}. Contemporary examples include Serbian (Latin and Cyrillic scripts), Malay (Latin and Jawi scripts), and Punjabi (Gurmukhi and Shahmukhi scripts). \citet{jung2023coexistence} further categorize digraphia into two types: complementary and exclusive. Japanese is an well-known example of complementary digraphia, where four scripts, Kanji, Hiragana, Katakana, and Latin letters are combined in common Japanese texts. In ASR, exclusive digraphia introduces multiple parallel written forms that can create ambiguities in audio transcription at the script level, while complementary digraphia can cause ambiguities at the orthography / spelling level~\cite{mazovetskiy24}.

For ASR, the definition of "writing system" extends beyond script to include orthography and spelling rules. Minor variations such as "color" vs. "colour" in American and British English can impact evaluations based on Word Error Rate (WER), where a spelling difference counts as a word error. While these variations have a limited effect in English due to their infrequency and can be addressed with rule-based normalization, languages with multiple coexisting spelling standards for substantial portions of their vocabulary will face significantly greater challenges, as discussed below.

\subsubsection{Norwegian Bokmål / Nynorsk}\label{sec:nbnn}

Norwegian exemplifies a digraphic language with two synchronic written standards: Bokmål (\texttt{nb\_no}) and Nynorsk (\texttt{nn\_no}). Table~\ref{tab:nbnnsentence} illustrates the differences with an example sentence. Notably, 4 out of 8 words differ in spelling, including common words like \textit{jeg} (Bokmål) and \textit{eg} (Nynorsk), both meaning "I" in English. Both MCV17 and FLEURS include Norwegian, but Common Voice uses \texttt{nn\_no} (purportedly Nynorsk) while FLEURS uses \texttt{nb\_no} (purportedly Bokmål). To verify the actual proportions of Bokmål and Nynorsk in these subsets, we employed a classification script detailed in Algorithm~\ref{nbnnscript}.

\begin{table}[ht]
\centering
\scalebox{0.8}{
\begin{tabular}{lp{7cm}}
English & Have I covered myself with song and playing the harp. \\\hline
Bokmål   & Har \textbf{eg dekt} meg med \textbf{song} og \textbf{harpespel}.\\
Nynorsk  & Har \textbf{jeg dekket} meg med \textbf{sang} og \textbf{harpespill}.           
\end{tabular}}
\caption{Spelling differences between Bokmål and Nynorsk. In this example, a mismatching orthography of the same sentence will lead to a 50\% WER even the transcription is completely "correct".}
\label{tab:nbnnsentence}
\end{table}

Table~\ref{tab:nbnnresult} presents the classification results. Neither MCV17 nor FLEURS contains purely Nynorsk or Bokmål as their language codes suggest. MCV17's \texttt{nn\_no} subset contains 8.1\% Bokmål sentences, while FLEURS' \texttt{nb\_no} subset contains 8.8\% Nynorsk. This finding suggests a possible lack of control over the sourcing of text sentences.

\begin{table}[]
\centering
\scalebox{0.85}{
\begin{tabular}{l|rr|rr}
                          & \multicolumn{2}{c|}{MCV17 \texttt{nn\_no}}      & \multicolumn{2}{c}{FLEURS \texttt{nb\_no}}  \\\hline
Nynorsk (\texttt{nn\_no}) & 764  & (65.1\%)             & 323   & (\textbf{8.8\%})    \\
Bokmål (\texttt{nb\_no})  & 96   & (\textbf{8.1\%})     & 2682  & (72.7\%)  \\
Mixed                     & 161  & (13.7\%)             & 344   & (9.3\%)    \\
Unmarked                  & 153  & (13.0\%)             & 338   & (9.1\%)    \\\hline
Total sentences           & 1174 & validated            & 3687  & \\
\end{tabular}}
\caption{Classification of Norwegian text prompts in MCV17 and FLEURS.}
\label{tab:nbnnresult}
\end{table}

We hypothesize that neglecting the Bokmål / Nynorsk distinction in an ASR dataset significantly affects model evaluation using WER. To test this, we evaluated a Conformer Hybrid Autoregressive Transducer (HAT) model \cite{variani20} of 120M parameters trained with Norwegian Bokmål data on both datasets. Table~\ref{tab:nower} shows the results. While the model exhibits similar deletion and insertion error rates across both datasets, the substitution error rate is nearly 25\% higher on MCV17's \texttt{nn\_no} subset. Manual inspection of the substitution errors revealed that most are indeed orthographic variants from Bokmål. This supports our hypothesis that mixing different writing systems in an ASR dataset significantly impacts downstream evaluations.

\begin{table}[]
\centering \scalebox{0.8}{
\begin{tabular}{p{0.8in}|p{1in}|p{1in}}
WER~[\%]         & MCV17 \texttt{nn\_no}      & FLEURS \texttt{nb\_no}\\ \hline
Total            & 49.1                       & 23.8  \\
Del/Ins/Sub      & 11.8 / 1.6 / \textbf{35.0} & 11.1 / 2.2 / \textbf{10.0}
\end{tabular}}
\caption{WER of a Norwegian Bokmål Conformer ASR model on MCV17 and FLEURS test splits. When tested on Nynorsk data, the substitution errors increase by 25\% abs.}
\label{tab:nower}
\end{table}

\subsubsection{Risks of Unverified Script Assumptions in Digraphic Languages}

Digraphia is not a static structure; it evolves, sometimes rapidly. The "default" script for a given language can shift, as is currently occurring in several post-Soviet countries \citep{jung2023coexistence}. For instance, Kazakhstan is transitioning from Cyrillic to Latin script by 2025, and Mongolia is restoring the Mongolian (Bichig) script by 2025 according to the government's plan. Neither MCV17 nor FLEURS include script codes in their locale codes, leading to implicit, unverified assumptions about the script used in the text data. These assumptions, listed in Table~\ref{tab:digraphicassumption} in Appendix, introduce significant risks for downstream applications, potentially compromising dataset usability in the near future.

\subsection{Ambiguous Register or Variety in Diglossic Languages}

Diglossia describes a community's use of two distinct language varieties in a compartmentalized manner: a "High" (H) variety for typically formal contexts and a "Low" (L) variety for everyday conversation, with the community perceiving these varieties as a single language \citep{ferguson1996epilogue}. This duality may lead to ambiguous language code interpretations in ASR dataset construction. For example, "I speak and write Chinese" could imply "I speak and write Mandarin", or "I speak Cantonese and write Mandarin". Such ambiguity can result in datasets pairing audio and text from mutually unintelligible varieties. Contemporary examples of diglossic languages include Standard Arabic (Fusha) and its vernacular dialects \citep{brustad2017diglossia, ferguson1996epilogue}, Hong Kong Chinese / Cantonese \citep{snow2010hong}, Standard German / Swiss German, Classical Tibetan / vernacular Tibetan \cite{roche2017introduction}, Persian \citep{mahmoodi201810} and Bengali \citep{dil1986diglossia}. Our analysis of MCV17 and FLEURS revealed serious issues with register and variant confusion, particularly in Arabic and Hong Kong Chinese / Cantonese. Since VoxPopuli does not include any diglossic languages, we find no such issues in VoxPopuli.

\subsubsection{Arabic}

Modern Standard Arabic (MSA, or Fusha) is a classical example of diglossia \citep{ferguson1959diglossia}, employed in formal contexts such as religion and education, while regional variants (e.g. Masri, Dārija) prevail in everyday conversation. These dialects can differ significantly from MSA \cite{hoigilt17}. We analyzed the Arabic text prompts in MCV17 (\texttt{ar}) and FLEURS (\texttt{ar\_eg}) using a classification tool (Algorithm \ref{al:ararbicscript} in Appendix). Table~\ref{tab:arabic} shows that both datasets mainly contain MSA, with some dialectal Arabic or mixed forms in MCV17's \texttt{ar} subset. Interestingly, FLEURS, labeled \texttt{ar\_eg} (presumably Egyptian Arabic), contains almost exclusively MSA. This could be due to a combination of factors: the intended dataset composition, the selection of source material, and the transcribers' interpretation of the term "Arabic", which can be ambiguous in a diglossic context. This highlights the complex interplay between dataset composition, content selection, and transcribers' preference and understanding of the task in diglossic situations.

\begin{table}[h!]
\centering \scalebox{0.85}{
\begin{tabular}{p{0.9in}|rr|rr}
          & \multicolumn{2}{c|}{MCV17 \texttt{ar}}  & \multicolumn{2}{c}{FLEURS \texttt{ar\_eg}}      \\ \hline
MSA (Fusha)         & \textbf{5963} & \textbf{(76.3\%)}   & \textbf{2787} &  \textbf{(98.6\%)} \\
Dialect       & 991           & (1.2\%)             & 0             &                     \\ 
Mixed         & 1648          & (2.1\%)             & 25            & (0.88\%) \\ 
Unmarked      & 15881         & (20.3\%)            & 15            & (0.53\%)  \\ \hline
\#sentences   & 78157         & validated           & 2827          &  \\ 
\end{tabular}}
\caption{Classification of text prompts in MCV17 \texttt{ar} and FLEURS \texttt{ar\_eg}. Nearly all data of FLEURS \texttt{ar\_eg} is in Fusha.}
\label{tab:arabic}
\end{table}

We speculate that the prevalence of MSA in FLEURS, despite the \texttt{ar\_eg} label, might be partly a consequence of adopting a strict language code formatting standard that does not adequately represent the nuances of diglossic languages. While regional codes exist within the ISO 639-3 standard, there is no widely accepted code for MSA within the ISO 3166-1 alpha-2 country code framework, which is often used in conjunction with language codes to form locale identifiers. The dataset creators might have chosen \texttt{eg} (Egypt) as the most readily available option to satisfy the IETF BCP 47 language tag specifications, despite it not accurately reflecting the linguistic reality of MSA as a supra-regional standard. This serves as a microcosm of a broader challenge: the limitations of current language classification systems (particularly within ISO 639-3) in representing regional, supra-regional and diglossic varieties, especially those that do not align neatly with national boundaries. For instance, there is no distinct code for African American Vernacular English (AAVE) or Scottish English within the current framework, hindering the ASR and TTS system development tailored to these communities. 


\subsubsection{Hong Kong Chinese / Cantonese} \label{sec:hkchinese}

Hong Kong presents another case of diglossia \citep{snow2010hong}, where Standard Written Chinese (SWC), a written register largely based on Modern Standard Mandarin, is used in formal writing and spoken Cantonese is used in everyday conversation. According to Unicode CLDR version 44.0,\footnote{\href{https://www.unicode.org/cldr/charts/44/supplemental/territory\_language\_information.html}{Territory-Language Information}} \texttt{zh} is interpreted as "Chinese", referring to SWC, while Cantonese has its own ISO 639-3 code, \texttt{yue}. However, the ambiguous nature of the code \texttt{zh} often leads to inconsistent interpretations in the industry.

We used the \texttt{canto-filter} package\footnote{\href{https://pypi.org/project/canto-filter/}{pypi.org/project/canto-filter/}} developed by \citet{lau-etal-2024-extraction} to classify text prompts into four categories. The results, shown in Table~\ref{table:cantonese}, highlight significant inconsistencies across datasets. Common Voice's \texttt{yue} subset aligns relatively well with Written Vernacular Cantonese (WVC), with most prompts being in WVC. However, Common Voice's \texttt{zh\_hk} subset contains a mixture of SWC and Cantonese, while FLEURS' \texttt{yue\_hk} subset consists almost entirely of SWC. This suggests that the \texttt{yue\_hk} label in FLEURS is likely a misnomer and should actually be \texttt{zh\_hk}.

\begin{table}[h!]
\centering
\renewcommand{\arraystretch}{1.1} 
\scalebox{0.7}{
\begin{tabular}{p{0.6in}|rr|rr|rr}
                & \multicolumn{2}{c|}{MCV17 \texttt{zh\_hk}}      & \multicolumn{2}{c|}{MCV17 \texttt{yue}}     & \multicolumn{2}{c}{FLEURS \texttt{yue\_hk}}  \\ \hline 
SWC             & \textbf{8851} & \textbf{(9.6\%)}        & 15     & (0.1\%)            & \textbf{2803} & \textbf{(89.8\%)}\\
Cantonese       & 37357         & (40.3\%)                & 16466  & (75.7\%)            & \textbf{0}    &          \\
Mixed           & 299           & (0.3\%)                 & 40     & (0.2\%)            & 0             &               \\
Unmarked        & 46113         & (49.8\%)                & 5238   & (24.1\%)            & 317           & (10.2\%)          \\ \hline
\#sentences     & 92620         &                         & 21759  &                     & 3120          &     \\
\end{tabular}}
\caption{Classification of text prompts in MCV17 \texttt{yue}, \texttt{zh\_hk}, and FLEURS \texttt{yue\_hk}. None of the FLEURS \texttt{yue\_hk} is Cantonese.}
\label{table:cantonese}
\end{table}

\subsection{Ambiguous Scoping of Target Dialect Continuum}

Both MCV17 and FLEURS utilize ISO 639-3 codes to identify languages. However, neither dataset specifies the dialect scope of each language subset in the metadata, leaving the interpretation of these codes open to contributors. Our investigation revealed that the \texttt{ff\_sn} (Fula, Senegal) subset in FLEURS only includes the Peul dialect spoken across Senegal. This is significant because the the Guinean variant of Fula has the most speakers, not the one found in the dataset. Another case we identify is the \texttt{kea\_cv} (Cape Verdean Creole, a.k.a. Kabuverdianu) in FLEURS, which consists entirely of the Sotavento (Southern Islands) variant. The omission of the Barlavento (Northern Islands) variant could limit the dataset's usefulness for developing speech systems that are robust across the entire dialect continuum of Cape Verdean Creole. 

The ambiguity in dialect scoping can skew the representativeness of the target language dataset as a whole. Researchers and developers might assume that a dataset labeled with a particular code encompasses the full range of variation within that language, when in reality it might only represent a limited subset of dialects. This can result in inflated performance metrics if ASR models are evaluated only on the dialects present in the training data, while performance on other dialects remains unknown and potentially lower~\cite{koenecke20,tang21,talafha24}. In TTS, having no control over the exact dialect might strike some sensitive nerve among native speakers, especially when the dialect boundaries align with historical and sociopolitical injustices.

\section{An Extreme Case Analysis: MCV17 \texttt{nan\_tw} (Taiwanese Southern Min)} \label{sec:extreme}

\subsection{Social and Historical Context}

Taiwanese Southern Min (TSM, a.k.a. Taiwanese Hokkien) has historically been an unwritten language. Various writing systems emerged sporadically, including full Sinograph (a.k.a. Chinese characters or Han script), Tâi-lô romanization, Church romanization (Pe̍h-ōe-jī), and a mixture of Sinographs and romanization \citep{ota2005investigation, alivin1999writing}. While the Taiwan Ministry of Education introduced the \textit{Taiwanese Southern Min Recommended Characters}\footnote{\href{https://language.moe.gov.tw/001/Upload/files/site_content/M0001/language_112/wp-content/uploads/2023/06/\%E6\%8E\%A8\%E8\%96\%A6\%E7\%94\%A8\%E5\%AD\%97700\%E5\%AD\%97.pdf}{\begin{CJK}{UTF8}{bsmi}臺灣閩南語推薦用字700字表\end{CJK}}} between 2007 and 2009, and a \textit{Dictionary of Frequently-Used Taiwan Minnan}\footnote{\href{https://sutian.moe.edu.tw/zh-hant/}{\begin{CJK}{UTF8}{bsmi}教育部臺灣台語常用詞辭典\end{CJK}}} to promote a standard orthography using full Sinographs, as of early 2025, the language community has not yet reached a consensus. As examples, a substantial proportion of pages of \href{https://zh-min-nan.wikipedia.org/wiki/Th\%C3\%A2u-ia\%CC\%8Dh}{Southern Min Wikipedia} use Latin letters, while everyday usage on social media platforms like Threads (Figure~\ref{fig:nan} in Appendix) displays a varied mix of Sinographs and romanizations. Unlike the more established complementary digraphia in Japanese, TSM lacks widely adopted rules for script choice, resulting in a more undetermined and dynamic case of complementary digraphia.

\subsection{Data Quality Issues}

Table~\ref{tab:nanexample} in Appendix presents a snapshot of the \texttt{nan\_tw} subset in MCV17, revealing critical issues that severely compromise its usability:

\begin{enumerate}
    \item \textbf{Dictionary Structure:} The dataset resembles a dictionary dump, with nearly every entry consisting of single words or short phrases, rather than complete sentences.
    \item \textbf{Duplicate Text Prompts:} Each prompt is written in both full Sinograph and full Latin (Tâi-lô), resulting in redundant entries.
    \item \textbf{Text-Audio Misalignment:} Voice contributors apparently read each prompt (single words or phrases) only once, leading to misalignment between the text (containing both writing systems) and the corresponding audio.
\end{enumerate}

Furthermore, many validated sentences (text prompts not yet paired with audio) are in \texttt{zh\_cn} (Mandarin Chinese in simplified characters), posing a significant risk of language contamination and potentially introducing the "mixed-language" issues discussed in Section \ref{sec:hkchinese}.

\subsection{Root Causes of the \texttt{nan\_tw} Data Issues}

To understand the rationale behind the dictionary-style structure and dual-script representation in MCV17 \texttt{nan\_tw}, we contacted the voice and text contributors. They explained that most TSM speakers are not proficient in reading or writing TSM. Moreover, among those who are literate in the language, preferences and proficiency levels for different writing standards vary considerably. Some are only comfortable with Sinographs, others with Romanization, while some prefer a mixture of both. To maximize participation and facilitate data collection, the contributors opted to include both Sinographs and Romanization, so that all potential contributors could read the prompts.

This situation highlights a common challenge faced by many low-resource or less-institutionalized languages when developing ASR datasets. ASR fundamentally involves transcribing audio, the spoken form, into text, the written form. But what if the language has no "written form"? The motivations of building ASR for such languages may differ fundamentally from those with well-established written traditions, necessitating distinct approaches to dataset construction. We will explore their implications in Section~\ref{sec:planning}.

\section{Discussions} \label{sec:impact}
\subsection{Impacts on Downstream Research and Applications} 

The quality issues discussed previously have significant ramifications for downstream research and applications. Shorter utterance lengths correlate with higher WER \citep{li2023asr, li2022fusing}, and excessive silence negatively impacts emotion recognition \citep{perez2022mind}. Lack of speaker and topic diversity can introduce biases related to gender, age, and regional accents \citep{feng2024towards, garnerin2021investigating}. While micro-level issues are often detectable programmatically, macro-level issues, stemming from sociolinguistic factors, can have more insidious and far-reaching impacts.

A compelling example is presented in \citet{costa2022no}, where a language identification (LangID) system, evaluated on the FLoRes-200 dataset (the text source for FLEURS), failed to distinguish between \texttt{zh\_hk} (Hong Kong Chinese) and \texttt{yue} (Cantonese), see the third confusion matrix in Figure 9 in the original paper. This stems exactly from the "wrong language" issue in FLEURS' \texttt{yue\_hk} subset, which predominantly contains SWC rather than Cantonese, as discussed in Section~\ref{sec:hkchinese} and publicly noted\footnote{\href{https://github.com/facebookresearch/flores/issues/61}{github.com/facebookresearch/flores/issues/61}}, and pointed out by \citet{lau-etal-2024-extraction}. Consequently, downstream models like Whisper-v3 struggle with these languages, exhibiting inconsistent outputs and unpredictable "auto-translations" between varieties.\footnote{\href{https://github.com/openai/whisper/discussions/366}{github.com/openai/whisper/discussions/366}} Furthermore, model distillation can exacerbate these issues, as seen in the WER degradation from 10.8\% to 46.1\% on Common Voice 15.0 \texttt{yue}.\footnote{\href{https://github.com/openai/whisper/discussions/2363}{github.com/openai/whisper/discussions/2363}}

This problem is covert because when training and test data are in the same SWC variety, the wrong output language might not be reflected by the WER, while downstream users expect the transcriptions to be in WVC consistently. The experiments from \citet{xie2025developing} further validates our findings, where the outputs of Whisper models are in the completely wrong register, and model performance significantly improved when fine-tuned with the pure Cantonese data. 

\subsection{Addressing Macro-Level Issues: The Role of Language Planning in Speech Dataset Creation} \label{sec:planning}

The macro-level issues identified in this study, unlike micro-level issues that can be mitigated programmatically (e.g. \citet{rai2024denoasr}), necessitate more fundamental improvements to dataset creation, particularly for less-institutionalized languages. We argue that it is crucial to incorporate language planning principles into the speech dataset design process.

Many widely spoken languages such as English, Spanish, Mandarin, are institutionalized and standardized due to a process called language planning \cite{cooper1989language}. This typically involves language planning agencies (LPAs), e.g. a country's education ministry, establishing a normative orthography, grammar, and lexicon to guide a diverse speech community. Consequently, speakers of such languages have good understanding of how to read and write them "correctly". However, with over 7,000 spoken languages and only around 200 countries in the world, most languages have not enjoyed such a privilege, leaving speakers without knowledge of how to read and write their mother tongues.

The decentralized, community-driven nature of Common Voice, while valuable for participation and diversity \citep{ardila2019common}, can exacerbate this issue. It can lead to implicit language planning decisions being made by communities without adequate expertise or consensus, as exemplified by the decision of merging Norwegian Nynorsk and Bokmål.\footnote{\href{https://discourse.mozilla.org/t/merging-norwegian-nynorsk-and-norwegian-bokmal/130474}{discourse.mozilla.org/t/merging-norwegian-nynorsk-and-norwegian-bokmal/130474}} As Common Voice expands to more languages with complex sociolinguistic backgrounds, such as Konkani,\footnote{\href{https://github.com/common-voice/common-voice/issues/4454}{github.com/common-voice/common-voice/issues/4454}} we anticipate a rise in these macro-level issues without proactive intervention.

Therefore, we propose the following checklist to follow when creating a new massively multilingual speech dataset, especially for less-institutionalized languages:

\begin{enumerate}
    \item \textbf{Sociolinguistic Assessment}: Before dataset creation, conduct a thorough sociolinguistic survey of the target language, including demographics, literacy rates, writing systems, diglossia/digraphia, language ideologies, and other relevant factors. 
    \item \textbf{Language Planning in Dataset Design}: If the language is not institutionalized (e.g. lacking a writing standard, written language is not vernacularized, multiple registers or scripts coexist), revisit the dataset design goal and clarify the objective. E.g. is the dataset in the vernacular or formal register, in Latin alphabet or the indigenous alphabet? Determine such design goals collaboratively with linguists, community members and native speakers to ensure downstream acceptance and practicality.
    \item \textbf{Proactive Prescription and Detailed Guidance}: Provide detailed transcription or recording guidelines to dataset contributors and annotators. Especially when the literacy rate is low or a standard orthography is lacking, prescribe specific guidelines on orthography, script and register choices to ensure annotators contribute in a consistent format.
    \item \textbf{Multi-Level Quality Assurance}: Implement rigorous quality assurance on both the text and audio. This includes checks based on automatic metrics (e.g., rejecting silent or extremely short audio, and text in wrong scripts) and human evaluation (e.g. rejecting text or audio in the wrong register, or in an out-of-scope dialect).
    \item \textbf{Comprehensive and Transparent Metadata}: Release the dataset with detailed metadata including the language planning decisions made in step 2. We recommend following the data statement practice by \citet{bender18}.
\end{enumerate}

When a language lacks a widely adopted written form, the motivation of developing ASR systems may shift from building a tech product to a tool for corpus planning and educational initiatives. As demonstrated by \citet{williams2013using} and \citet{kumar2012improving}, ASR can be employed to reduce illiteracy and promote the use of a standard written form within a community. 

Previous research has pointed out the effectiveness of social network and media technologies in codifying polycentric orthography standards \cite{leggio201713}. As various groups have been reportedly using Mozilla Common Voice to democratize speech technology for their mother tongues,\footnote{\href{https://www.technologyreview.com/2024/11/15/1106935/how-this-grassroots-effort-could-make-ai-voices-more-diverse}{How this grassroots effort could make AI voices more diverse.}} we believe such community-driven projects can take the critical role of LPAs in establishing a written norm within the language community \cite{sackett2017community, mccarty2018community}.  

\section{Conclusions} \label{sec:conclusions}

This study investigated the quality of prominent public speech datasets, highlighting the critical need for sociolinguistic awareness and the importance of language planning in dataset design, especially for less-institutionalized languages. We also proposed actionable guidelines for future speech dataset creation.

For downstream users of existing datasets, we strongly recommend human evaluation by native speakers or linguists, and data cleaning/filtering before use. Precise scoping of language/dialect variants is crucial to avoid mixed-language issues in training multilingual models. We also recommend employing more flexible evaluation metrics than WER and CER, such as those proposed by \citet{nigmatulina-etal-2020-asr} and \citet{karita2023lenient}.

Future work can focus on developing practical tools and frameworks for implementing our proposed guidelines. We also recommend closely monitoring Common Voice's ongoing expansion into new languages. Future research should also address the limitations of current language classification systems, e.g. the ISO 639-3 standard, in representing the full spectrum of linguistic diversity and nuanced sociolinguistic realities.

\subsection{Speech Dataset as a tool for Language Planning}

The creation of crowdsourced datasets like Common Voice offers a unique microcosm of community-based language planning in action. This process inherently surfaces and forces negotiation on unresolved questions of orthography, lexicon, and dialectal norms. We argue that this phenomenon should not be viewed as an obstacle, but as a critical opportunity. Such dataset construction, particularly for less-institutionalized languages, should be re-conceptualized as not merely data collection, but a live, community-led language planning initiative. As \citet{markl-mcnulty-2022-language} argue,  ASR datasets are vital "infrastructures" for language communities, but their maintenance and curation are often lacking.

We argue that a critical next step for the research community is to explore how to deliberately leverage this process. Learning from \citet{10.1145/3613904.3642026} who enables speech technologies even for unwritten languages, future work should focus on developing methodologies and frameworks that empower communities to use dataset creation as an explicit tool for grassroots language planning and revitalization.

\section{Limitations}


While this study covered over 40 languages, a significant number remain uninspected in the evaluated datasets. Future work should extend our investigation to these languages and develop practical tools and frameworks for implementing the proposed guidelines. Our proposed framework assumes access to linguistic expert and native speaker resource. Further research can focus on how small teams and community can overcome the lack of such resource and build sociolinguistically-informed datasets.

\section{Ethical Considerations}

There are no known ethical concerns or risks associated with this work.

\bibliography{custom}

\clearpage
\appendix \label{sec:appendix}

\onecolumn

\section{License or Terms for Use of Artifacts}

In this paper we studied three datasets VoxPopuli, FLEURS and MCV17. VoxPopuli is under CC-BY-NC 4.0 license\footnote{\href{https://github.com/facebookresearch/voxpopuli/blob/main/LICENSE}{github.com/facebookresearch/voxpopuli/blob/main/LICENSE}}, FLEURS is under CC-BY 4.0 license\footnote{\href{https://huggingface.co/datasets/google/fleurs}{huggingface.co/datasets/google/fleurs}} and MCV17 is under CC0 1.0 license\footnote{\href{https://huggingface.co/datasets/mozilla-foundation/common_voice_17_0}{huggingface.co/datasets/mozilla-foundation/common\_voice\_17\_0}}. In addition, we used canto-filter package in \ref{sec:hkchinese}. The package is under MIT license\footnote{\href{https://pypi.org/project/canto-filter/}{pypi.org/project/canto-filter/}}. The usage of each dataset and package in our work is consistent with its intended use according to the license.

\section{Meta-Information of VoxPopuli, FLEURS and MCV17}
\label{sec:datasetdetail}

The VoxPopuli corpus contains 1.8k hours of transcribed speech in 16 European languages, detailed in Table \ref{tab:vpstats}.

\begin{table*}[h]
\centering

  \begin{tabular}{llr|llr}
  code & name & hours & code & name & hours \\ \hline
\texttt{cs} & Czech & 62 & \texttt{hu} & Hungarian & 63 \\		
\texttt{de} & German & 282 & \texttt{it} & Italian & 91 \\		
\texttt{en} & English & 543 & \texttt{lt} & Lithuanian & 2 \\		
\texttt{es} & Spanish & 166 & \texttt{nl} & Dutch & 53 \\		
\texttt{et} & Estonian & 3 & \texttt{pl} & Polish & 111 \\		
\texttt{fi} & Finnish & 27 & \texttt{ro} & Romanian & 89 \\		
\texttt{fr} & French & 211 & \texttt{sk} & Slovak & 35 \\		
\texttt{hr} & Croatian & 43 & \texttt{sl} & Slovene & 10 \\		
  \end{tabular}
  \caption{List of languages and hours in VoxPopuli}
  \label{tab:vpstats}
\end{table*}

FLEURS contains English utterances their translations and readings into 101 languages. Around 2009 English sentences are extracted from FLoRes101 corpus. Each sentence is recorded by 3 native speakers and the invalid recordings are discarded, which makes a total of 1.4k hours of speech and aroudn 12 hours in each language. The full list of languages is shown in Table \ref{tab:fleursstats1} and \ref{tab:fleursstats2}.

\begin{table*}[h]
\centering
  \begin{tabular}{ll|ll}
  code & name & code & name \\ \hline
\texttt{af\_za} & Afrikaans, South Africa & \texttt{el\_gr} & Greek, Greece \\
\texttt{am\_et} & Amharic, Ethiopia & \texttt{en\_us} & English, United States \\
\texttt{ar\_eg} & Arabic, Egypt & \texttt{es\_419} & Spanish, Latin America \\
\texttt{as\_in} & Assamese, India & \texttt{et\_ee} & Estonian, Estonia \\
\texttt{ast\_es} & Asturian, Spain & \texttt{fa\_ir} & Persian, Iran \\
\texttt{az\_az} & Azerbaijani, Azerbaijan & \texttt{ff\_sn} & Fulah, Senegal \\
\texttt{be\_by} & Belarusian, Belarus & \texttt{fi\_fi} & Finnish, Finland \\
\texttt{bg\_bg} & Bulgarian, Bulgaria & \texttt{fil\_ph} & Filipino, Philippines \\
\texttt{bn\_in} & Bengali, India & \texttt{fr\_fr} & French, France \\
\texttt{bs\_ba} & Bosnian, Bosnia & \texttt{ga\_ie} & Irish, Ireland \\
\texttt{ca\_es} & Catalan, Spain & \texttt{gl\_es} & Galician, Spain \\
\texttt{ceb\_ph} & Cebuano, Philippines & \texttt{gu\_in} & Gujarati, India \\
\texttt{ckb\_iq} & Central Kurdish, Iraq & \texttt{ha\_ng} & Hausa, Nigeria \\
\texttt{cmn\_hans\_cn} & Mandarin, China & \texttt{he\_il} & Hebrew, Israel \\
\texttt{cs\_cz} & Czech, Czech Republic & \texttt{hi\_in} & Hindi, India \\
\texttt{cy\_gb} & Welsh, United Kingdom & \texttt{hr\_hr} & Croatian, Croatia \\
\texttt{da\_dk} & Danish, Denmark & \texttt{hu\_hu} & Hungarian, Hungary \\
\texttt{de\_de} & German, Germany & \texttt{hy\_am} & Armenian, Armenia \\
  \end{tabular}
  \caption{List of languages in FLEURS, Part 1}
  \label{tab:fleursstats1}
\end{table*}

\clearpage

\begin{table*}[h]
\centering
  \begin{tabular}{ll|ll}
  code & name & code & name \\ \hline
\texttt{id\_id} & Indonesian, Indonesia & \texttt{ny\_mw} & Chichewa, Malawi \\
\texttt{ig\_ng} & Igbo, Nigeria & \texttt{oc\_fr} & Occitan, France \\
\texttt{is\_is} & Icelandic, Iceland & \texttt{om\_et} & Oromo, Ethiopia \\
\texttt{it\_it} & Italian, Italy & \texttt{or\_in} & Odia, India \\
\texttt{ja\_jp} & Japanese, Japan & \texttt{pa\_in} & Punjabi, India \\
\texttt{jv\_id} & Javanese, Indonesia & \texttt{pl\_pl} & Polish, Poland \\
\texttt{ka\_ge} & Georgian, Georgia & \texttt{ps\_af} & Pashto, Afghanistan \\
\texttt{kam\_ke} & Kamba, Kenya & \texttt{pt\_br} & Portuguese, Brazil \\
\texttt{kea\_cv} & Kabuverdianu, Cape Verde & \texttt{ro\_ro} & Romanian, Romania \\
\texttt{kk\_kz} & Kazakh, Kazakhstan & \texttt{ru\_ru} & Russian, Russia \\
\texttt{km\_kh} & Khmer, Cambodia & \texttt{sd\_in} & Sindhi, India \\
\texttt{kn\_in} & Kannada, India & \texttt{sk\_sk} & Slovak, Slovakia \\
\texttt{ko\_kr} & Korean, South Korea & \texttt{sl\_si} & Slovenian, Slovenia \\
\texttt{ky\_kg} & Kyrgyz, Kyrgyzstan & \texttt{sn\_zw} & Shona, Zimbabwe \\
\texttt{lb\_lu} & Luxembourgish, Luxembourg & \texttt{so\_so} & Somali, Somalia \\
\texttt{lg\_ug} & Ganda, Uganda & \texttt{sr\_rs} & Serbian, Serbia \\
\texttt{ln\_cd} & Lingala, DRC & \texttt{sv\_se} & Swedish, Sweden \\
\texttt{lo\_la} & Lao, Laos & \texttt{sw\_ke} & Swahili, Kenya \\
\texttt{lt\_lt} & Lithuanian, Lithuania & \texttt{ta\_in} & Tamil, India \\
\texttt{luo\_ke} & Luo, Kenya & \texttt{te\_in} & Telugu, India \\
\texttt{lv\_lv} & Latvian, Latvia & \texttt{tg\_tj} & Tajik, Tajikistan \\
\texttt{mi\_nz} & Māori, New Zealand & \texttt{th\_th} & Thai, Thailand \\
\texttt{mk\_mk} & Macedonian, North Macedonia & \texttt{tr\_tr} & Turkish, Turkey \\
\texttt{ml\_in} & Malayalam, India & \texttt{uk\_ua} & Ukrainian, Ukraine \\
\texttt{mn\_mn} & Mongolian, Mongolia & \texttt{umb\_ao} & Umbundu, Angola \\
\texttt{mr\_in} & Marathi, India & \texttt{ur\_pk} & Urdu, Pakistan \\
\texttt{ms\_my} & Malay, Malaysia & \texttt{uz\_uz} & Uzbek, Uzbekistan \\
\texttt{mt\_mt} & Maltese, Malta & \texttt{vi\_vn} & Vietnamese, Vietnam \\
\texttt{my\_mm} & Burmese, Myanmar & \texttt{wo\_sn} & Wolof, Senegal \\
\texttt{nb\_no} & Norwegian Bokmål, Norway & \texttt{xh\_za} & Xhosa, South Africa \\
\texttt{ne\_np} & Nepali, Nepal & \texttt{yo\_ng} & Yoruba, Nigeria \\
\texttt{nl\_nl} & Dutch, Netherlands & \texttt{yue\_hant\_hk} & Cantonese, Hong Kong \\
\texttt{nso\_za} & Northern Sotho, South Africa & \texttt{zu\_za} & Zulu, South Africa \\
  \end{tabular}
  \caption{List of languages in FLEURS, Part 2}
  \label{tab:fleursstats2}
\end{table*}

\clearpage

The statistics of MCV17 meta-information can be found in Table \ref{tab:mcvstats1} and \ref{tab:mcvstats2}. The full dataset contains 124 locales. Each locale contains 3 categories: validated, invalidated and other. Validated means the data clip has received more than 2 validations and the upvotes $>$ downvotes; Invalidated means the data clip has received more than 2 validations but the upvotes $\leq$ downvotes; Other means the data clip has not received 2 or more validations.

When analyzing the speaker diversity in Section \ref{speakerdiversity}, we used all the 3 categories. In other analyses, we only used the validated data and skipped locales with 0 or 1 utterances.

\begin{table*}[h]
\centering

  \begin{tabular}{lp{0.7in}>{\raggedleft\arraybackslash}p{0.4in}>{\raggedleft\arraybackslash}p{0.5in}|lp{0.7in}>{\raggedleft\arraybackslash}p{0.4in}>{\raggedleft\arraybackslash}p{0.5in}}
code & name & total hours & validated hours & code & name & total hours & validated hours \\ \hline
\texttt{ab} & Abkhaz & 84.41 & 59.86 & \texttt{es} & Spanish & 2219.39 & 561.82 \\
\texttt{af} & Afrikaans & 0.56 & 0.27 & \texttt{et} & Estonian & 59.97 & 45.85 \\
\texttt{am} & Amharic & 2.63 & 1.55 & \texttt{eu} & Basque & 672.41 & 273.48 \\
\texttt{ar} & Arabic & 155.81 & 90.27 & \texttt{fa} & Persian & 415.36 & 363.42 \\
\texttt{as} & Assamese & 3.28 & 2.72 & \texttt{fi} & Finnish & 21.65 & 13.35 \\
\texttt{ast} & Asturian & 1.91 & 0.81 & \texttt{fr} & French & 1147.41 & 1013.03 \\
\texttt{az} & Azerbaijani & 0.45 & 0.19 & \texttt{fy\_nl} & West Frisian & 212.09 & 68.73 \\
\texttt{ba} & Bashkir & 268.17 & 257.77 & \texttt{ga\_ie} & Irish & 10.47 & 5.48 \\
\texttt{bas} & Basaa & 2.82 & 2.16 & \texttt{gl} & Galician & 131.47 & 65.57 \\
\texttt{be} & Belarusian & 1765.25 & 1717.14 & \texttt{gn} & Guarani & 27.56 & 3.66 \\
\texttt{bg} & Bulgarian & 20.77 & 16.46 & \texttt{ha} & Hausa & 12.17 & 3.94 \\
\texttt{bn} & Bengali & 1272.95 & 53.51 & \texttt{he} & Hebrew & 6.06 & 2.22 \\
\texttt{br} & Breton & 26.87 & 18.55 & \texttt{hi} & Hindi & 20.7 & 14.11 \\
\texttt{ca} & Catalan & 3586.54 & 2697.77 & \texttt{hsb} & Upper Sorbian & 3.02 & 2.43 \\
\texttt{ckb} & Central Kurdish & 172.16 & 130.21 & \texttt{ht} & Haitian Creole & 0.01 & 0 \\
\texttt{cnh} & Chin Haka & 6.04 & 2.4 & \texttt{hu} & Hungarian & 172.63 & 92.64 \\
\texttt{cs} & Czech & 262.67 & 76.09 & \texttt{hy\_am} & Armenian & 47.36 & 22.27 \\
\texttt{cv} & Chuvash & 27.51 & 24.36 & \texttt{ia} & Interlingua & 17.04 & 13.73 \\
\texttt{cy} & Welsh & 156.58 & 123.08 & \texttt{id} & Indonesian & 64.53 & 28.93 \\
\texttt{da} & Danish & 12.6 & 11.68 & \texttt{ig} & Igbo & 8.77 & 0.02 \\
\texttt{de} & German & 1423.9 & 1333.93 & \texttt{is} & Icelandic & 0.08 & 0.02 \\
\texttt{dv} & Dhivehi & 64.21 & 38.87 & \texttt{it} & Italian & 395.57 & 354.96 \\
\texttt{dyu} & Dyula & 0.49 & 0.32 & \texttt{ja} & Japanese & 476.98 & 124.31 \\
\texttt{el} & Greek & 31.5 & 18.64 & \texttt{ka} & Georgian & 214.21 & 138.8 \\
\texttt{en} & English & 3507.22 & 2614.76 & \texttt{kab} & Kabyle & 689.77 & 566.48 \\
\texttt{eo} & Esperanto & 1904.97 & 1433.1 & \texttt{kk} & Kazakh & 3.46 & 2.13 \\
\texttt{es} & Spanish & 2219.39 & 561.82 & \texttt{kmr} & Kurmanji Kurdish & 99.79 & 67.57 \\
\texttt{et} & Estonian & 59.97 & 45.85 & \texttt{ko} & Korean & 5.57 & 1.72 \\
\end {tabular}
\caption{List of locales and hours in Common Voice 17.0, Part 1}
\label{tab:mcvstats1}
\end{table*}

\clearpage

\begin{table*}[h]
  \begin{tabular}{lp{0.7in}>{\raggedleft\arraybackslash}p{0.4in}>{\raggedleft\arraybackslash}p{0.5in}|lp{0.7in}>{\raggedleft\arraybackslash}p{0.4in}>{\raggedleft\arraybackslash}p{0.5in}}
code & name & total hours & validated hours & code & name & total hours & validated hours \\ \hline
\texttt{ky} & Kyrgyz & 47.6 & 38.42 & \texttt{sah} & Sakha & 12.65 & 8.31 \\
\texttt{lg} & Ganda & 559.22 & 436.7 & \texttt{sat} & Santali & 1.02 & 0.57 \\
\texttt{lij} & Ligurian & 3.29 & 2.8 & \texttt{sc} & Sardinian & 1.95 & 1.5 \\
\texttt{lo} & Lao & 0.37 & 0.2 & \texttt{sk} & Slovak & 26.89 & 22.1 \\
\texttt{lt} & Lithuanian & 25.21 & 23.71 & \texttt{skr} & Saraiki & 6.59 & 4.2 \\
\texttt{ltg} & Latgalian & 23.3 & 20.67 & \texttt{sl} & Slovenian & 14.99 & 11.38 \\
\texttt{lv} & Latvian & 276.61 & 222.45 & \texttt{sq} & Albanian & 1.97 & 1.94 \\
\texttt{mdf} & Moksha & 0.5 & 0.49 & \texttt{sr} & Serbian & 6.76 & 5.01 \\
\texttt{mhr} & Meadow Mari & 301.25 & 280.45 & \texttt{sv\_se} & Swedish & 54.5 & 45.38 \\
\texttt{mk} & Macedonian & 22.89 & 7.82 & \texttt{sw} & Swahili & 1084.72 & 399.48 \\
\texttt{ml} & Malayalam & 10.11 & 3.46 & \texttt{ta} & Tamil & 403.84 & 232.59 \\
\texttt{mn} & Mongolian & 23.08 & 13.17 & \texttt{te} & Telugu & 2.3 & 0.26 \\
\texttt{mr} & Marathi & 27.48 & 18.75 & \texttt{th} & Thai & 422.96 & 171.29 \\
\texttt{mrj} & Western Mari & 36.77 & 33.63 & \texttt{ti} & Tigrinya & 0.11 & 0.03 \\
\texttt{mt} & Maltese & 17.21 & 8.48 & \texttt{tig} & Tigre & 2.69 & 1.08 \\
\texttt{myv} & Erzya & 3.2 & 3.15 & \texttt{tk} & Turkmen & 6.52 & 2.73 \\
\texttt{nan\_tw} & Southern Min (Taiwan) & 20.21 & 5.73 & \texttt{tok} & Toki Pona & 18.86 & 13.59 \\
\texttt{ne\_np} & Nepali & 1.54 & 0.81 & \texttt{tr} & Turkish & 122.06 & 117.28 \\
\texttt{nhi} & Nahuatl & 0.03 & 0.02 & \texttt{tt} & Tatar & 31.24 & 30.59 \\
\texttt{nl} & Dutch & 119.6 & 109.48 & \texttt{tw} & Twi & 0.27 & 0.16 \\
\texttt{nn\_no} & Norwegian Nynorsk & 1.67 & 1.42 & \texttt{ug} & Uyghur & 236.63 & 199.46 \\
\texttt{nso} & Northern Sotho & 0.03 & 0 & \texttt{uk} & Ukrainian & 111.71 & 97.44 \\
\texttt{oc} & Occitan & 12.83 & 2.25 & \texttt{ur} & Urdu & 231.8 & 63.52 \\
\texttt{or} & Odia & 12.51 & 4.4 & \texttt{uz} & Uzbek & 263.24 & 99.63 \\
\texttt{os} & Ossetic & 0.31 & 0.28 & \texttt{vi} & Vietnamese & 18.76 & 5.65 \\
\texttt{pa\_in} & Punjabi & 3.99 & 2.01 & \texttt{vot} & Votic & 0.29 & 0.06 \\
\texttt{pl} & Polish & 175.85 & 166.71 & \texttt{yi} & Yiddish & 0.05 & 0.04 \\
\texttt{ps} & Pashto & 2.11 & 1.69 & \texttt{yo} & Yoruba & 7.29 & 5.07 \\
\texttt{pt} & Portuguese & 210.97 & 174.17 & \texttt{yue} & Cantonese & 177.76 & 23.4 \\
\texttt{quy} & Quechua (Ayacucho) & 0.01 & 0 & \texttt{zgh} & Tamazight & 1.47 & 0.51 \\
\texttt{rm\_sursilv} & Romansh Sursilvan & 10.91 & 6.53 & \texttt{zh\_cn} & Chinese (China) & 1061.32 & 233.77 \\
\texttt{rm\_vallader} & Romansh Vallader & 4.26 & 2.46 & \texttt{zh\_hk} & Chinese (Hong Kong) & 138.23 & 107.42 \\
\texttt{ro} & Romanian & 46.8 & 19.85 & \texttt{zh\_tw} & Chinese (Taiwan) & 126.04 & 77.1 \\
\texttt{ru} & Russian & 273.89 & 234.46 & \texttt{zu} & Zulu & 0.05 & 0 \\
\texttt{rw} & Kinyarwanda & 2384 & 2001.34 & \texttt{zza} & Zaza & 0.45 & 0.33 \\
\end {tabular}
\caption{List of locales and hours in Common Voice 17.0, Part 2}
\label{tab:mcvstats2}
\end{table*}

\clearpage

\section{Median Utterance Duration of VoxPopuli, FLEURS and MCV17} \label{sec:medianduration}

\begin{figure*}[h]
\centering
  \includegraphics[width=0.95\linewidth]{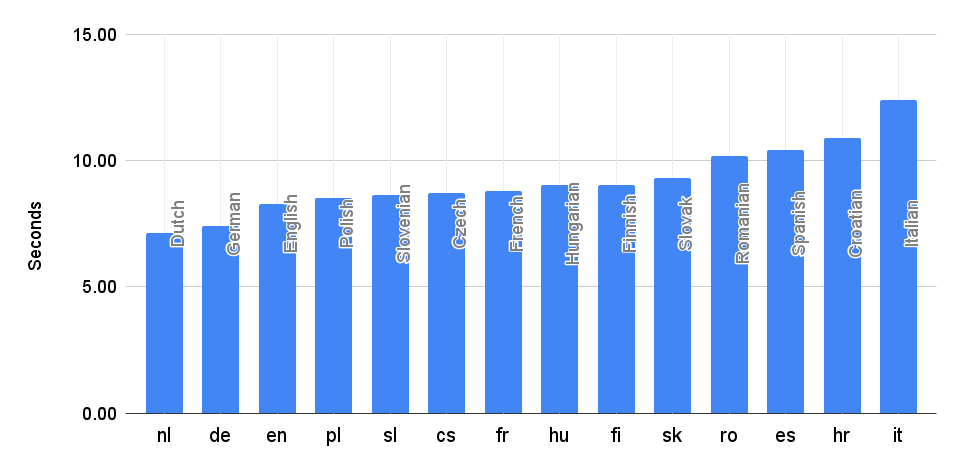}
  \caption{Median utterance duration of the 14 languages in VoxPopuli. All languages have a median utterance duration of at least 7 seconds.}
  \label{fig:vpduration}
\end{figure*}

\begin{figure*}[h]
\centering
  \includegraphics[width=1\linewidth]{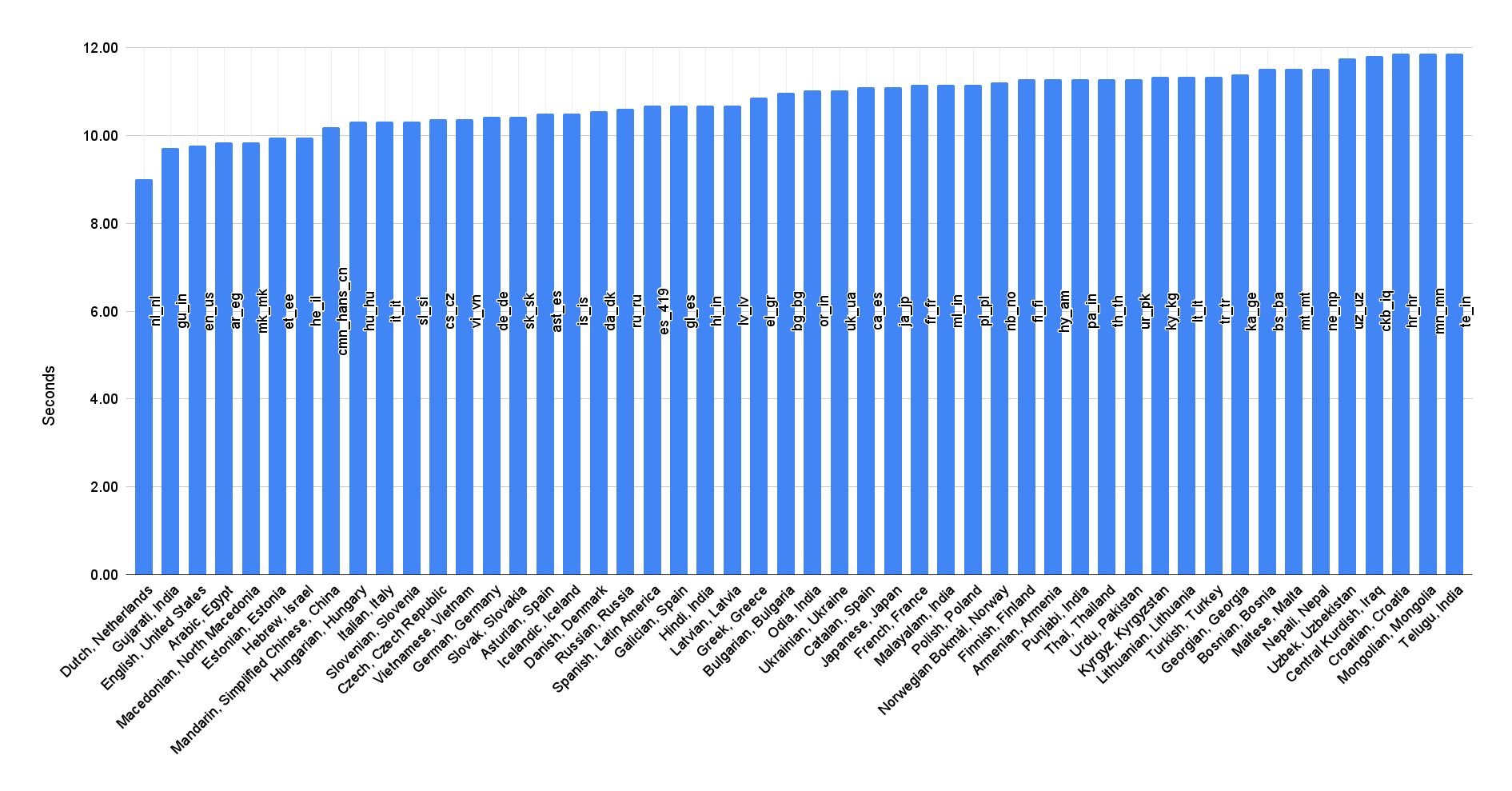}
  \caption{50 languages with the shortest median utterance duration in FLEURS. All languages have a median utterance duration of at least 5 seconds.}
  \label{fig:fleursduration}
\end{figure*}

\begin{figure*}[h]
\centering
  \includegraphics[width=1\linewidth]{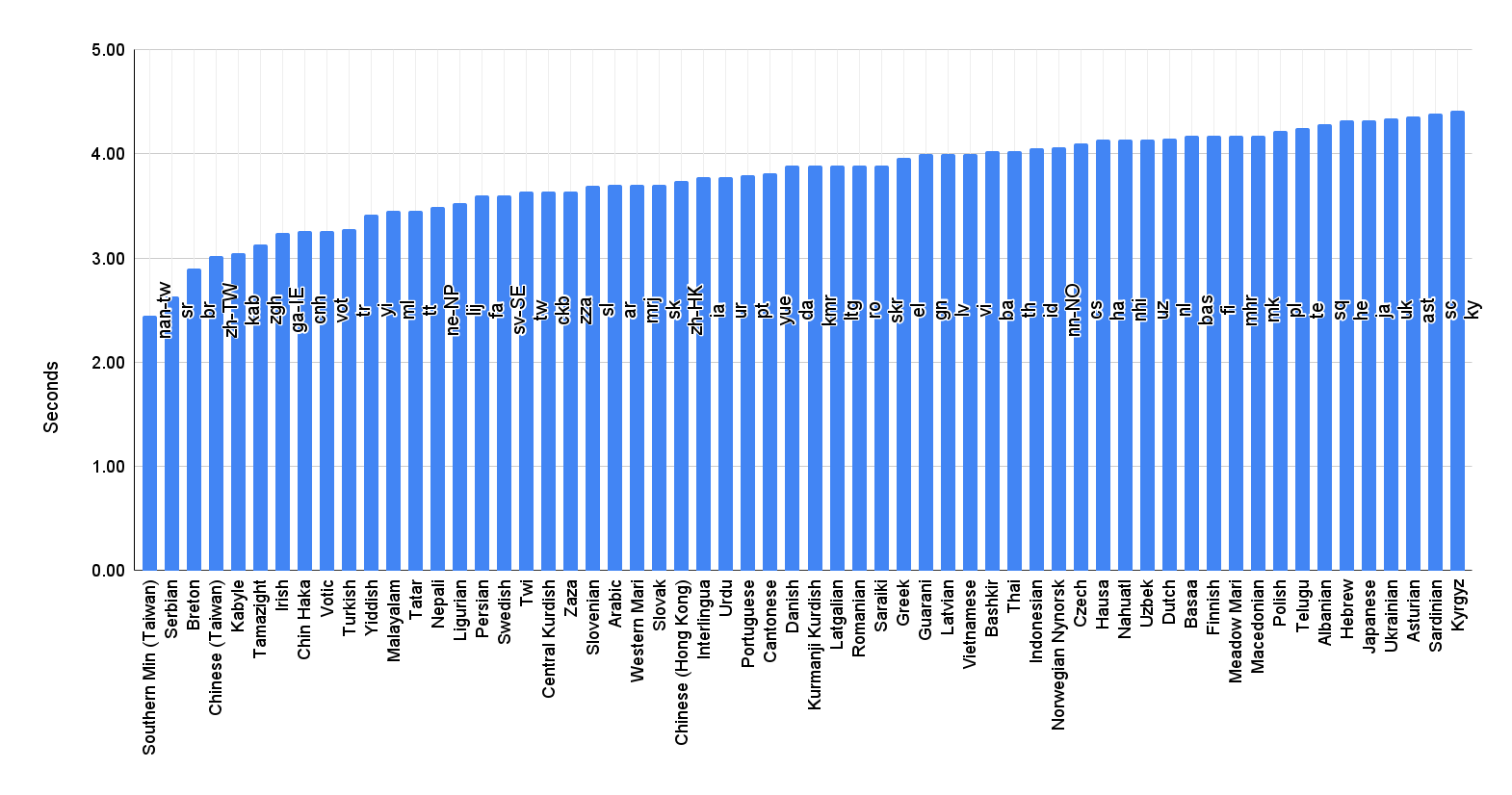}
  \caption{60 languages with the shortest median utterance duration in Mozilla Common Voice 17.0. Taiwanese Southern Min (\texttt{nan\_tw}) has a median duration of only 2.45 seconds.}
  \label{fig:mcv17duration}
\end{figure*}

\begin{figure*}[h]
\centering
  \includegraphics[width=1\linewidth]{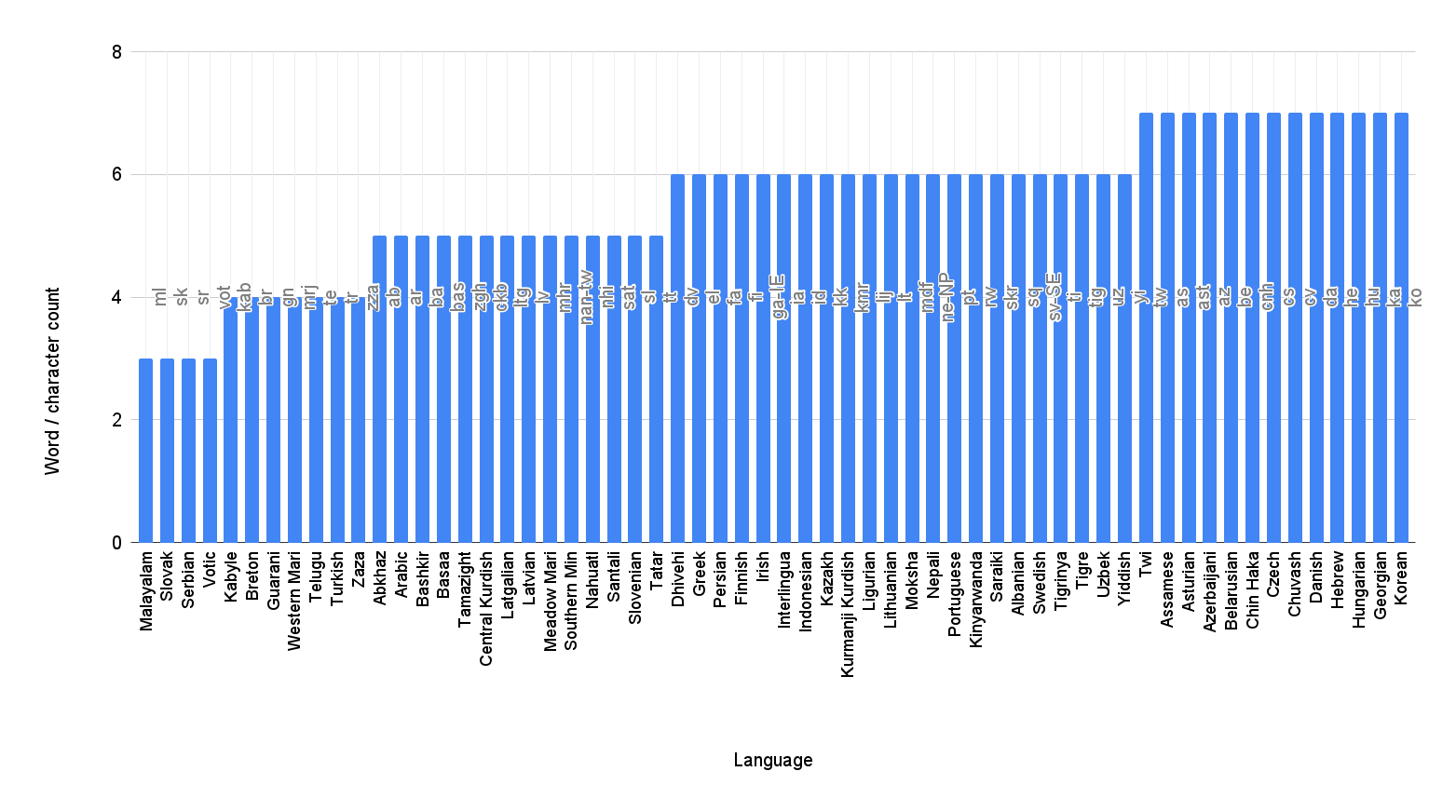}
  \caption{60 languages with the shortest median text prompt length (measured by word or character count) in Mozilla Common Voice 17.0. Note that \texttt{nan\_tw} doesn't have the shortest length because the words are duplicated in two writing systems.}
  \label{fig:mcv17len}
\end{figure*}

\clearpage
\section{Speech and Silence Percentage of VoxPopuli, FLEURS and MCV17} \label{sec:speechsilence}

\begin{figure*}[h]
\centering
  \includegraphics[width=1\linewidth]{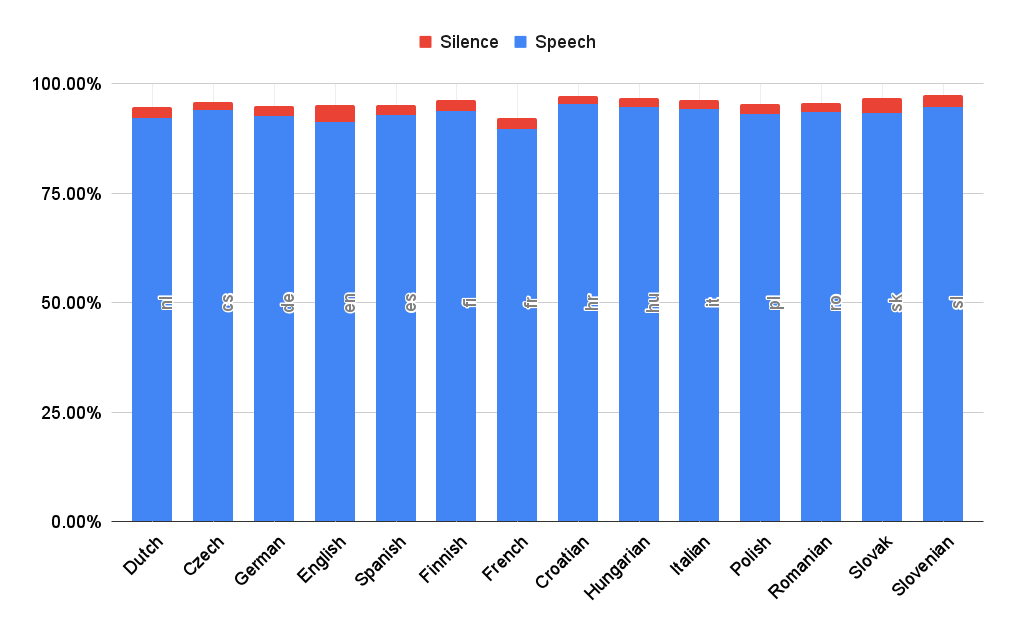}
  \caption{Percentages of speech and silence in VoxPopuli. All languages have more than 90\% speech in the utterances. Note that the percentages of silence and speech don't add up to 100\% because we omit other types of sound such as music, noise, laughter, etc.}
  \label{fig:vpsilence}
\end{figure*}

\begin{figure*}[h]
\centering
  \includegraphics[width=1\linewidth]{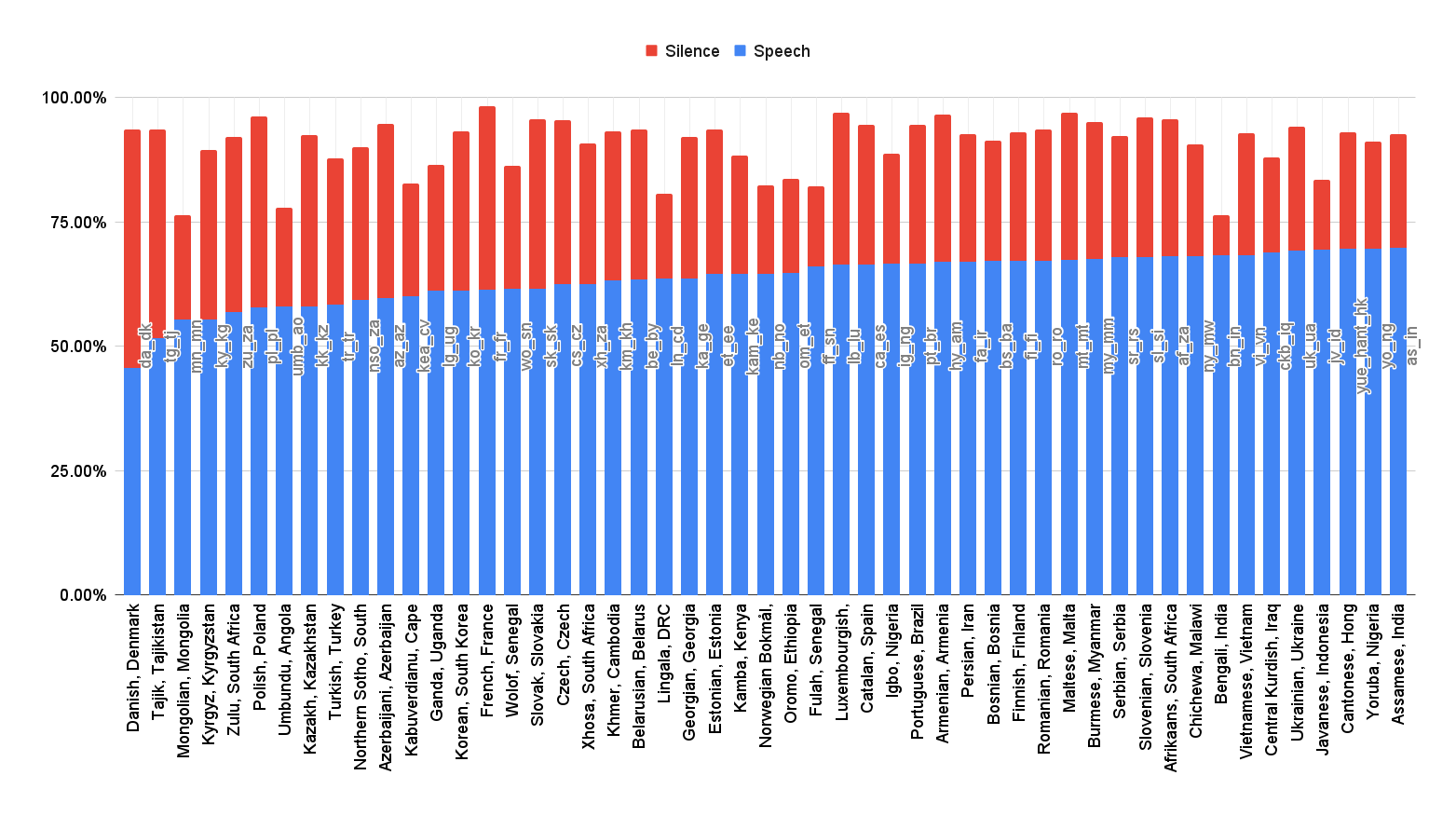}
  \caption{50 languages with the lowest speech proportion in FLEURS. Danish (\texttt{da\_dk}) has less than 50\% speech in the audio recordings. Note that the percentages of silence and speech don't add up to 100\% because we omit other types of sound such as music, noise, laughter, etc.}
  \label{fig:fleurssilence}
\end{figure*}

\begin{figure*}[h]
\centering
  \includegraphics[width=1\linewidth]{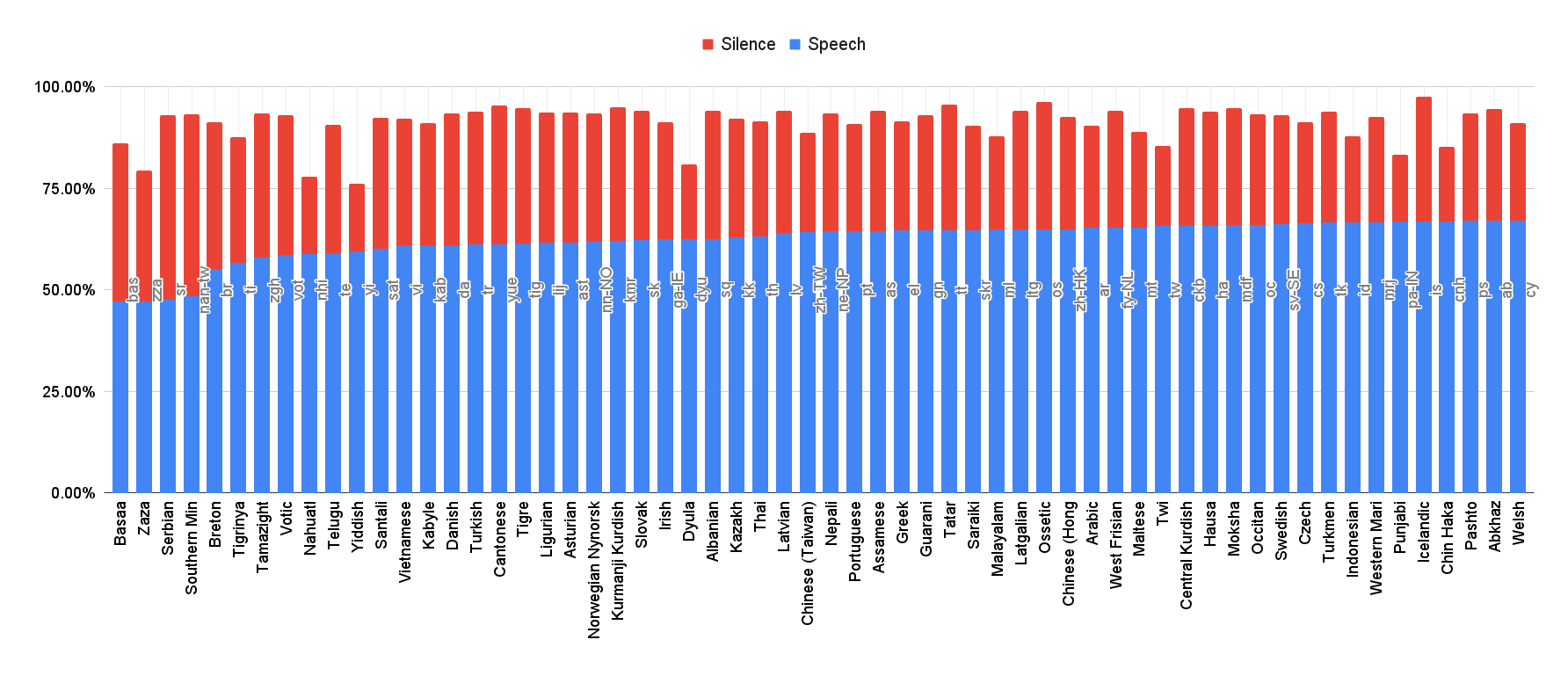}
  \caption{60 languages with the lowest speech proportion in Mozilla Common Voice 17.0. Basaa, Zaza, Serbian and Taiwanese Southern Min have less than 50\% speech in the audio recordings. Note that the percentages of silence and speech don't add up to 100\% because we omit other types of sound such as music, noise, laughter, etc.}
  \label{fig:mcv17silence}
\end{figure*}

\clearpage
\section{Example Sentences of MCV17} \label{examplesentences}

\begin{table}[ht]
\centering
\bgroup
\def\arraystretch{1.2}
\scalebox{1}{
\begin{tabular}{p{6.5cm}|p{4cm}}
\texttt{zh\_cn} (Chinese, China) & \texttt{ca} (Catalan) \\ \hline
\begin{CJK}{UTF8}{gbsn}殿试登进士第三甲第一百一十一名。\end{CJK} & No he anat mai a Agost. \\
\begin{CJK}{UTF8}{gbsn}殿试登进士第三甲第一百七十一名。\end{CJK} & No he anat mai a Aigües. \\
\begin{CJK}{UTF8}{gbsn}殿试登进士第三甲第一百三十一名。\end{CJK} & No he anat mai a Aiora. \\
\begin{CJK}{UTF8}{gbsn}殿试登进士第三甲第一百三十九名。\end{CJK} & No he anat mai a Aiòder. \\
\begin{CJK}{UTF8}{gbsn}殿试登进士第三甲第一百三十五名。\end{CJK} & No he anat mai a Alaior. \\ 
\begin{CJK}{UTF8}{gbsn}殿试登进士第三甲第一百九十名。\end{CJK} & No he anat mai a Alaró. \\ 
\begin{CJK}{UTF8}{gbsn}殿试登进士第三甲第一百五十名。\end{CJK} & No he anat mai a Alaior. \\ 
\begin{CJK}{UTF8}{gbsn}殿试登进士第三甲第一百八十七名。\end{CJK} & No he anat mai a Albaida. \\
\begin{CJK}{UTF8}{gbsn}殿试登进士第三甲第一百八十九名。\end{CJK} & No he anat mai a Albatera.\\
\begin{CJK}{UTF8}{gbsn}殿试登进士第三甲第一百六十七名。\end{CJK} & No he anat mai a Alberic. \\
... &  ... \\
\end{tabular}}
\egroup
\caption{\label{tab:repeat}Examples of highly repetitive, seemingly machine-generated sentences in Common Voice 17.0. We find at least hundreds of such sentences in \texttt{zh\_cn}, \texttt{zh\_tw}, \texttt{zh\_hk} and \texttt{ca}. We suspect that similar issues may exist in other languages of MCV17 that we haven't inspected.}
\end{table}

\begin{table}[ht]
\centering \scalebox{1}{
\begin{tabular}{l}
\texttt{nan\_tw} sentences in MCV17 \\ \hline
\begin{CJK}{UTF8}{bsmi}竹仔籃（Tik-á-nâ）\end{CJK} \\
\begin{CJK}{UTF8}{bsmi}竹南鎮（Tek-lâm-tìn）\end{CJK} \\
\begin{CJK}{UTF8}{bsmi}竹南鎮（Tik-lâm-tìn）\end{CJK} \\
\begin{CJK}{UTF8}{bsmi}竹坑口（Tik-khinn-kháu | Tek-khiⁿ-kháu） \end{CJK}\\
\begin{CJK}{UTF8}{bsmi}竹子腳（Tek-á-kha）\end{CJK} \\
\begin{CJK}{UTF8}{bsmi}竹崎鄉（Tik-kiā-hiong）\end{CJK}\\
\begin{CJK}{UTF8}{bsmi}竹東鎮（Tik-tang-tìn）\end{CJK} \\
\begin{CJK}{UTF8}{bsmi}竹東（Tik-tang）	\end{CJK} \\
\begin{CJK}{UTF8}{bsmi}竹田鄉（Tik-tshân-hiong）\end{CJK} \\
\begin{CJK}{UTF8}{bsmi}竹田（Tik-tshân）\end{CJK} \\
...
\end{tabular}}
\caption{A snapshot of MCV17 \texttt{nan\_tw}'s text prompts. Each row is a single word or phrase with its romanization. For homographs, multiple pronunciations are appended to the Sinographs.}
\label{tab:nanexample}
\end{table}

\clearpage
\section{Average Hours per Speaker of MCV17} 

\begin{table}[ht]
  \centering
  \scalebox{0.9}{
  \begin{tabular}{p{1in}p{1.5in}@{}rrr}
  code  &  name         &     unique   &         total     &  average  hours \\
        &               &  speakers    &         hours     &  per speaker \\ \hline
 \texttt{rw}   &  Kinyarwanda  &     1131     &         2,384.0  &       2.11     \\
 \texttt{mk}   &  Macedonian   &     19       &         22.9     &       1.20     \\
 \texttt{eo}   &  Esperanto    &     1739     &         1,905.0  &       1.10     \\
 \texttt{lg}   &  Ganda        &     657      &         559.2    &       0.85     \\
 \texttt{sw}   &  Swahili      &     1452     &         1,084.7  &       0.75     \\
 \texttt{ur}   &  Urdu         &     349      &         231.8    &       0.66     \\
 \texttt{mrj}   &  Western~Mari &    60        &         36.8     &       0.61  \\
 \texttt{mhr}   &  Meadow~Mari  &    496       &         301.3    &       0.61  \\
 \texttt{kab}   &  Kabyle       &     1547     &         689.8    &       0.45     \\
 \texttt{ta}   &  Tamil        &     906      &         403.8    &       0.45     \\
 \texttt{mr}   &  Marathi      &     90       &         27.5     &       0.31     \\
 \texttt{ha}   &  Hausa        &     40       &         12.2     &       0.30     \\
 \texttt{ba}   &  Bashkir      &     917      &         268.2    &       0.29     \\
 \texttt{he}   &  Hebrew       &     21       &         6.1      &       0.29     \\
 \texttt{cs}   &  Czech        &     983      &         262.7    &       0.27     \\
 \texttt{ia}   &  Interlingua  &     67       &         17.0     &       0.25     \\
 \texttt{myv}   &  Erzya        &     13       &         3.2      &       0.25     \\
 \texttt{cv}   &  Chuvash      &     112      &         27.5     &       0.25     \\
 \texttt{ps}   &  Pashto       &     9        &         2.1      &       0.23     \\
 \texttt{be}& Belarusian & 8291 & 1,765.25 & 0.21 \\
\texttt{ab}& Abkhaz & 403 & 84.41 & 0.21 \\
\texttt{yue} & Cantonese & 913 & 177.76 & 0.19 \\
\texttt{ug}& Uyghur & 1258 & 236.63 & 0.19 \\
\texttt{dv}& Dhivehi & 357 & 64.21 & 0.18 \\
\texttt{kmr} & Kurmanji Kurdish & 561 & 99.79 & 0.18 \\
\texttt{lij} & Ligurian & 19 & 3.29 & 0.17 \\
\texttt{ky}& Kyrgyz & 283 & 47.60 & 0.17 \\
\texttt{gn}& Guarani & 164 & 27.56 & 0.17 \\
\texttt{bg}& Bulgarian & 134 & 20.77 & 0.15 \\
\texttt{zh\_cn} & Chinese (China) & 7005 & 1,061.32 & 0.15 \\
\texttt{hsb} & Upper Sorbian & 21 & 3.02 & 0.14 \\
\texttt{sc}& Sardinian & 14 & 1.95 & 0.14 \\
\texttt{br}& Breton & 207 & 26.87 & 0.13 \\
\texttt{ka}& Georgian & 1679 & 214.21 & 0.13 \\
\texttt{tok} & Toki Pona & 149 & 18.86 & 0.13 \\
\texttt{hy\_an} & Armenian & 390 & 47.36 & 0.12 \\
\texttt{uz}& Uzbek & 2170 & 263.24 & 0.12 \\
\texttt{rm\_sursilv} & Romansh Sursilvan & 90 & 10.91 & 0.12 \\
\texttt{tt}& Tatar & 258 & 31.24 & 0.12 \\
  \end{tabular}}
  \caption{Top 40 languages with the highest average total hours per speaker in Common Voice 17.0. As a reference, English has 0.04 hours per speaker.}
  \label{tab:hoursperspeaker}
  \end{table}

\begin{table*}[h]
  \begin{tabular}{lp{0.68in}>{\raggedleft\arraybackslash}p{0.3in}>{\raggedleft\arraybackslash}p{0.4in}>{\raggedleft\arraybackslash}p{0.4in}|lp{0.55in}>{\raggedleft\arraybackslash}p{0.3in}>{\raggedleft\arraybackslash}p{0.4in}>{\raggedleft\arraybackslash}p{0.4in}}
code & name & unique speakers & total hours & average hours per speaker & code & name & unique speakers & total hours & average hours per speaker \\ \hline
\texttt{zu} & Zulu & 1 & 0.05 & 0.05 & \texttt{bas} & Basaa & 36 & 2.82 & 0.08 \\
\texttt{nso} & Northern Sotho & 1 & 0.03 & 0.03 & \texttt{nn\_no} & Norwegian Nynorsk & 38 & 1.67 & 0.04 \\
\texttt{ht} & Haitian Creole & 1 & 0.01 & 0.01 & \texttt{te} & Telugu & 39 & 2.30 & 0.06 \\
\texttt{nhi} & Nahuatl & 2 & 0.03 & 0.02 & \texttt{ha}& Hausa & 40 & 12.17 & 0.30 \\
\texttt{quy} & Quechua (Ayacucho) & 2 & 0.01 & 0.01 & \texttt{as} & Assamese & 46 & 3.28 & 0.07 \\
\texttt{yi} & Yiddish & 3 & 0.05 & 0.02 & \texttt{rm\_vallader} & Romansh Vallader & 53 & 4.26 & 0.08 \\
\texttt{zza} & Zaza & 4 & 0.45 & 0.11 & \texttt{sq}& Albanian & 55 & 1.97 & 0.04 \\
\texttt{is} & Icelandic & 4 & 0.08 & 0.02 & \texttt{skr} & Saraiki & 57 & 6.59 & 0.12 \\
\texttt{vot} & Votic & 6 & 0.29 & 0.05 & \texttt{mrj} & Western Mari & 60 & 36.77 & 0.61 \\
\texttt{tw} & Twi & 6 & 0.27 & 0.05 & \texttt{ia} & Interlingua & 67 & 17.04 & 0.25 \\
\texttt{ti} & Tigrinya & 6 & 0.11 & 0.02 & \texttt{pa\_in} & Punjabi & 68 & 3.99 & 0.06 \\
\texttt{os} & Ossetic & 8 & 0.31 & 0.04 & \texttt{mr} & Marathi & 90 & 27.48 & 0.31 \\
\texttt{ps} & Pashto & 9 & 2.11 & 0.23 & \texttt{rm\_sursilv} & Romansh Sursilvan & 90 & 10.91 & 0.12 \\
\texttt{mdf} & Moksha & 11 & 0.50 & 0.05 & \texttt{ko} & Korean & 90 & 5.57 & 0.06 \\
\texttt{myv} & Erzya & 13 & 3.20 & 0.25 & \texttt{yo} & Yoruba & 108 & 7.29 & 0.07 \\
\texttt{sat} & Santali & 13 & 1.02 & 0.08 & \texttt{sah} & Sakha & 111 & 12.65 & 0.11 \\
\texttt{lo} & Lao & 13 & 0.37 & 0.03 & \texttt{cv} & Chuvash & 112 & 27.51 & 0.25 \\
\texttt{sc} & Sardinian & 14 & 1.95 & 0.14 & \texttt{tk} & Turkmen & 112 & 6.52 & 0.06 \\
\texttt{zgh} & Tamazight & 17 & 1.47 & 0.09 & \texttt{ig} & Igbo & 114 & 8.77 & 0.08 \\
\texttt{mk} & Macedonian & 19 & 22.89 & 1.20 & \texttt{or} & Odia & 125 & 12.51 & 0.10 \\
\texttt{lij} & Ligurian & 19 & 3.29 & 0.17 & \texttt{bg} & Bulgarian & 134 & 20.77 & 0.15 \\
\texttt{he} & Hebrew & 21 & 6.06 & 0.29 & \texttt{ml} & Malayalam & 134 & 10.11 & 0.08 \\
\texttt{hsb} & Upper Sorbian & 21 & 3.02 & 0.14 & \texttt{oc} & Occitan & 145 & 12.83 & 0.09 \\
\texttt{af} & Afrikaans & 23 & 0.56 & 0.02 & \texttt{tok} & Toki Pona & 149 & 18.86 & 0.13 \\
\texttt{tig} & Tigre & 24 & 2.69 & 0.11 & \texttt{sr} & Serbian & 153 & 6.76 & 0.04 \\
\texttt{az} & Azerbaijani & 26 & 0.45 & 0.02 & \texttt{sl} & Slovenian & 154 & 14.99 & 0.10 \\
\texttt{ast} & Asturian & 29 & 1.91 & 0.07 & \texttt{gn} & Guarani & 164 & 27.56 & 0.17 \\
\texttt{am} & Amharic & 30 & 2.63 & 0.09 & \texttt{kk} & Kazakh & 166 & 3.46 & 0.02 \\
\texttt{ne}\_np & Nepali & 32 & 1.54 & 0.05 & \texttt{ga\_ie} & Irish & 192 & 10.47 & 0.05 \\
\texttt{dyu} & Dyula & 33 & 0.49 & 0.01 & \texttt{br} & Breton & 207 & 26.87 & 0.13
  \end{tabular}
  \caption{Bottom 60 languages with the least unique voice contributors in Common Voice 17.0. As a reference, English has 92325 speakers, with each contributing 0.04 hours on average.}
  \label{tab:speakers}
  \end{table*}


\clearpage
\section{Script Choices of Digraphic Languages in MCV17 and FLEURS}
\begin{table*}[h]
    \begingroup
    \renewcommand{\arraystretch}{1.4} 
    \scalebox{0.8}{
    \begin{tabular}{ll|ll}
        Language name & Scripts that can be written in  & MCV17   & FLEURS  \\ 
        \hline
        Central Kurdish (Sorani) &  Cyrillic, Hawar (Latin), Sorani (Arabic)   &   \texttt{ckb}: Arabic & \texttt{ckb\_iq}: Sorani (Arabic) \\
        Dyula & Latin, N’Ko         & \texttt{dyu}: Latin    & N/A      \\
        Fula & Adlam, Ajami (Arabic), Latin & N/A & \texttt{ff\_sn}: Latin \\
        Malay & Jawi (Arabic), Latin & \texttt{ms}: Latin    & \texttt{ms\_my}: Latin    \\
        Mongolian & Cyrillic, Mongolian (Bichig)  & \texttt{mn}: Cyrillic  & \texttt{mn\_mn}: Cyrillic  \\
        Northern Kurdish (Kurmanji) & Cyrillic, Hawar (Latin), Sorani (Arabic) & \texttt{kmr}: Hawar (Latin) & N/A\\ 
        Serbian  & Cyrillic, Latin  & \texttt{sr}: Cyrillic   & \texttt{sr\_rs}: Cyrillic and Latin \\
        Punjabi  & Gurmukhi, Shahmukhi (Arabic) & \texttt{pa\_in}: Gurmukhi & \texttt{pa\_in}: Gurmukhi  \\
        Tamazight (Berber a.k.a. Amazigh) & Arabic, Latin, Tifinagh  & \texttt{zgh}: Tifinagh &    N/A\\
        Uzbek    & Arabic, Cyrillic, Latin & \texttt{uz}: Latin   & \texttt{uz\_uz}: Latin  \\
        Votic    & Cyrillic, Latin  & \texttt{vot}: Latin  & N/A  \\
    \end{tabular}}
    \endgroup
\caption{Script choices and assumptions of digraphic / trigraphic languages in Common Voice 17.0 and FLEURS.}
\label{tab:digraphicassumption}
\end{table*}

\section{Example of Mixed-script Writings in Taiwanese Southern Min}
\begin{figure}[ht]
  \includegraphics[width=\columnwidth]{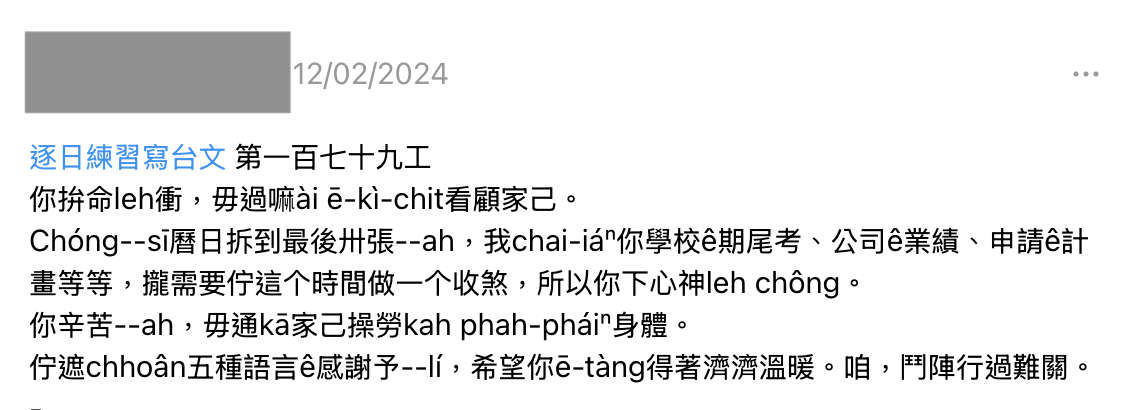}
  \caption{A Threads post written in Taiwanese Southern Min. Sinographs and Latin letters are used interchangeably without a codified convention.}
  \label{fig:nan}
\end{figure}

\clearpage
\section{Text Classification Logic}

\begin{algorithm*}
\begin{algorithmic}[1]
\Procedure{ClassifyBokmalNynorsk}{$sentence$}
    
    \State $nynorskMarkers \gets$ ``ikkje'', ``eg'', ``eit'', ``eitt'', ``me'', ``ho'', ``hjå'', ``kva'', ``kven'', ``noko'', ``nokre'', ``sjå'', ``skule'', ``kor'', ``fyrst'', ``mykje'', ``òg'', ``medan''
    \State $bokmaalMarkers \gets$ ``ikke'', ``jeg'', ``et'', ``en'', ``vi'', ``hun'', ``hos'', ``hva'', ``hvem'', ``noe'', ``noen'', ``se'', ``skole'', ``hvor'', ``først'', ``mye'', ``også'', ``mens''
    
    \State $sentence \gets$ Convert $sentence$ to lowercase
    
    \State $nynorskCount \gets 0$
    \State $bokmaalCount \gets 0$
    
    \For{$marker \in nynorskMarkers$}
        \If{$marker$ exists as whole word in $sentence$}
            \State $nynorskCount \gets nynorskCount + 1$
        \EndIf
    \EndFor
    
    \For{$marker \in bokmaalMarkers$}
        \If{$marker$ exists as whole word in $sentence$}
            \State $bokmaalCount \gets bokmaalCount + 1$
        \EndIf
    \EndFor
    
    \State $nynorskCount \gets nynorskCount + $ Count of words ending in 'a'
    \State $bokmaalCount \gets bokmaalCount + $ Count of words ending in 'en'
    
    \If{$nynorskCount > bokmaalCount$}
        \State \Return ``Nynorsk''
    \ElsIf{$bokmaalCount > nynorskCount$}
        \State \Return ``Bokmål''
    \ElsIf{$nynorskCount = bokmaalCount$ and $nynorskCount > 0$}
        \State \Return ``Mixed''
    \Else
        \State \Return ``Unmarked / Neutral''
    \EndIf
\EndProcedure
\end{algorithmic}
\caption{Norwegian Orthography Classification Logic}
\label{nbnnscript}
\end{algorithm*}

\begin{algorithm*}
\begin{algorithmic}[1]
\Procedure{IsFusha}{$sentence$}
    \State $fushaMarkers \gets $ List of Fusha (Modern Standard Arabic) markers 
    \State $dialectMarkers \gets $ List of dialectal Arabic markers
    
    \State $fushaScore \gets 0$
    \State $dialectScore \gets 0$
    
    \For{each $marker$ in $fushaMarkers$}
        \If{$marker$ found in $sentence$}
            \State $fushaScore \gets fushaScore + 1$
        \EndIf
    \EndFor
    
    \For{each $marker$ in $dialectMarkers$}
        \If{$marker$ found in $sentence$}
            \State $dialectScore \gets dialectScore + 1$
        \EndIf
    \EndFor

    \If{$fushaScore = 0$ and $dialectScore = 0$}
        \State \Return "unmarked"
    \ElsIf{$fushaScore > dialectScore$}
        \State \Return "fusha"
    \ElsIf{$fushaScore < dialectScore$}
        \State \Return "dialect"
    \Else
        \State \Return "mixed"
    \EndIf
\EndProcedure
\end{algorithmic}
\caption{Arabic Fusha / Dialect Classification Logic}
\label{al:ararbicscript}
\end{algorithm*}

\clearpage
\section{Languages Inspected by Native Speaker Volunteers in this Study}

We asked native speaker volunteers from our co-workers to review 100 randomly sampled sentences (text and audio) for coherence, audio-text alignment, dialect, topic domain, and language ID from each language subsets of the datasets. The language list is shown in \ref{tab:qual}.

\begin{table}[h]
  \begingroup
  \renewcommand{\arraystretch}{1.4} 
  \scalebox{0.8}{
  \begin{tabular}{l|ll|l|ll}
  Name & MCV17 code & FLEURS code & Name & MCV17 code & FLEURS code \\ \hline
  Amharic & \texttt{am} & \texttt{am\_et} & Lingala &  & \texttt{ln\_cd} \\
  Arabic (Egypt) & \texttt{ar} & \texttt{ar\_eg} & Mandarin Chinese (China) & \texttt{zh\_cn} & \texttt{cmn\_hans\_cn} \\
  Basaa & \texttt{bas} &  & Mandarin Chinese (Taiwan) & \texttt{zh\_tw} &  \\
  Bashkir & \texttt{ba} &  & Meadow Mari & \texttt{mhr} &  \\
  Cantonese & \texttt{yue} & \texttt{yue\_hk} & Moksha & \texttt{mdf} &  \\
  Cape Verde Creole &  & \texttt{kea\_cv} & Norwegian (Nynorsk / Bokmål) & \texttt{nn\_no} & \texttt{nb\_no} \\
  Catalan & \texttt{ca} & \texttt{ca\_es} & Oromo &  & \texttt{om\_et} \\
  Chichewa &  & \texttt{ny\_mw} & Persian & \texttt{fa} &  \texttt{fa\_ir}\\
  Chinese (Hong Kong) & \texttt{zh\_hk} &  & Russian & \texttt{ru} & \texttt{ru\_ru} \\
  Chuvash & \texttt{cv} &  & Shona &  & \texttt{sn\_zw} \\
  Danish, Denmark & \texttt{da} & \texttt{da\_dk} & Southern Min & \texttt{nan\_tw} &  \\
  Dyula & \texttt{dyu} &  & Spanish & \texttt{es} &  \\
  English (United States) & \texttt{en} & \texttt{en\_us} & Swahili & \texttt{sw} & \texttt{sw\_ke} \\
  Fulah (Senegal) &  & \texttt{ff\_sn} & Twi & \texttt{tw} &  \\
  Ganda & \texttt{lg} & \texttt{lg\_ug} & Ukrainian & \texttt{uk} &  \\
  German & \texttt{de} &  & Votic & \texttt{vot} &  \\
  Greek & \texttt{el} & \texttt{el\_gr} & Western Mari & \texttt{mrj} &  \\
  Hausa &  & \texttt{ha\_ng} & Wolof &  & \texttt{wo\_sn} \\
  Igbo &  & \texttt{ig\_ng} & Yoruba &  & \texttt{yo\_ng} \\
  Japanese & \texttt{ja} & \texttt{ja\_jp} & Zulu & \texttt{zu} & \texttt{zu\_za} \\
  Korean & \texttt{ko} & \texttt{ko\_kr} & & & 
  \end{tabular}}
  \endgroup
  \caption{Languages with native speaker volunteers available for qualitative inspections in this study.}
  \label{tab:qual}
  \end{table}
  
\section{Acknowledgments}

The authors gratefully acknowledge the helpful comments from Daan van Esch, Sandy Ritchie, and the consults by Jordan Sihno Mbeleg, Fafi Getachew, Issagha Ba, Ali Konaté, Geraldine Nalubeg, Zacharie Liman-Tinguiri, Chike Ezeokoli, Chuck Onwuzuru, Hernani Delgado Chan, Anya Mhone, Senzeni Mpofu, Isaac Ackah, Isaac Attuah, Temi Adesina, Fatou Fall, Michael Asare, Temi Adesina, Marvin Ngobeni and Parisa Haghani.

\end{document}